# Performance assessment and exhaustive listing of 500+ nature-inspired metaheuristic algorithms

Zhongqiang Ma[a], Guohua Wu[a,*], P.N. Suganthan[b,c,*], Aijuan Song[a], Qizhang Luo[a]

[a]School of Traffic and Transportation Engineering, Central South University, Changsha 410075, China

[b]KINDI Center for Computing Research, College of Engineering, Qatar University, Doha, Qatar

[c]School of Electrical and Electronic Engineering, Nanyang Technological University, 639798, Singapore

**Abstract:** Metaheuristics are popularly used in various fields, and they have attracted much attention in the scientific and industrial communities. In recent years, the number of new metaheuristic names has been continuously growing. Generally, the inventors attribute the novelties of these new algorithms to inspirations from either biology, human behaviors, physics, or other phenomena. In addition, these new algorithms, compared against basic versions of other metaheuristics using classical benchmark problems without shift/rotation, show competitive performances. However, many new metaheuristics are not rigorously tested on challenging benchmark suites and are not compared with state-of-the-art metaheuristic variants. Therefore, in this study, we exhaustively tabulate more than 500 metaheuristics. In particular, several representative metaheuristics are introduced from two aspects, namely, the inspirational source and the essential operators for generating solutions. To comparatively evaluate the performance of the recent competitive variants and newly proposed metaheuristics, 11 newly proposed metaheuristics (generally with high numbers of citations) and 4 variants of established metaheuristics are comprehensively compared on the CEC2017 benchmark suite. For fair comparisons, a parameter tuning tool named irace is used to automatically configure the parameters of all 15 algorithms. In addition, whether these algorithms have a search bias to the origin (i.e., the center of the search space) is investigated. All the experimental results are analyzed by several nonparametric statistical methods, including the Bayesian rank-sum test, Friedman test, Wilcoxon signed-rank test, critical difference plot and Bayesian signed-rank test. Moreover, the convergence, diversity, and the trade-off between exploration and exploitation of these 15 algorithms are also analyzed. The results show that the performance of the newly proposed EBCM (effective butterfly optimizer with covariance matrix adaptation) algorithm performs comparably to the 4 well performing variants of the established metaheuristics and possesses similar properties and behaviors, such as convergence, diversity, exploration and exploitation trade-offs, in many aspects. However, the other 10 recent metaheuristics are less efficient and robust than the 4 recent variants of the established metaheuristics. The performance of all 15 of the algorithms is likely to deteriorate due to certain transformations, while the 4 state-of-the-art metaheuristics are less affected by transformations such as the shifting of the global optimal point away from the center of the search space. It should be noted that, except EBCM, the other 10 new algorithms proposed mostly during 2019-2020 are inferior to the well performing 2017 variants of differential evolution and evolution strategy in terms of convergence speed and global search ability on CEC 2017 functions. Moreover, the other 10 new algorithms are rougher (i.e., behave with high oscillations) in terms of the trade-off between exploitation and exploration and population diversity compared with the 4 state-of-the-art algorithms. Finally, several important issues relevant to the metaheuristic research area are discussed and some potential research directions are suggested.

uation, nonparametric tests

## 1 Introduction

Optimization algorithms play an important role in the economy, engineering, management, and medicine because many real-world problems can be modeled as optimization problems. Optimization algorithms attempt to reach the optimal objective values (i.e., minimum or maximum) and satisfy the related constraints. Very complex problems are highly constrained, multimodal, discontinuous, noisy and of high dimension, all of which can make the traditional exact algorithms (e.g., mathematical programming) ineffective.

As an alternative method, approximate algorithms have attracted much attention in recent decades. Approximate algorithms can be roughly divided into heuristic algorithms and metaheuristic algorithms. Heuristic algorithms generally need to be elaborately designed for specific optimization problems and may have weak flexibility in solving other types of problems. In contrast, metaheuristics provide a general optimization framework for solving various optimization problems and benefit from the randomness embedded into the operators, which makes it possible to find a satisfactory, or near-optimal solution, in a reasonable time, however, they cannot guarantee the optimum solution for complex problems [1]. The merits of simplicity, less problem dependence, flexibility, derivative-free mechanism, and local optima avoidance make metaheuristics user-friendly [2].

Metaheuristics can be defined as high-level methodologies that embody the underlying heuristics to solve optimization problems [3]. The term metaheuristic was first proposed by Glover in 1986 [4], and most modern nature-inspired algorithms can be considered metaheuristics [5]. The concept of nature-inspired is about creating algorithms by mimicking natural phenomena or biological behaviors to solve optimization problems. For example, simulated annealing (SA) [6] is inspired by the idea of the solid annealing principle. Particle swarm optimization (PSO) [7] is derived from the interaction behaviors of birds in the flock. Ant colony optimization (ACO) [8] mimics the behaviors of ants in finding the shortest path between a nest and a food source. The classification criteria of metaheuristics can be varied. For instance, according to the number of candidate solutions at each iteration, metaheuristics can be further divided into population-solution based metaheuristics and single-solution based metaheuristics [9]. Popular single-solution based metaheuristics include SA, tabu search (TS) [4], iterated local search (ITS) [10], guided local search (GLS) [11], random search (RS) [12], variable neighborhood search (VNS) [13], and large neighborhood search (LNS) [14]. Population-solution based metaheuristics include the genetic algorithm (GA) [15], differential evolution (DE) [16], pattern search (PS) [17], and others.

There are still some issues in the field of metaheuristics. With the increase in the number of recent metaheuristics, the necessity of irrationally introducing new metaheuristic algorithms is questioned [18]. Molina et al. [19] found that there is no necessary significant relationship between the inspiration sources of algorithms and their performance. However, some researchers expect to improve the performance of metaheuristics through the inspiration source, which is still misleading. There is no work that comprehensively evaluates and compares the efficiency and effectiveness of the newly proposed and the state-of-the-art metaheuristics [20,21]. Furthermore, some algorithms perform well on problems with the optimal solution located at the origin (i.e., center of the search space) but are less efficient when the optimal solutions are shifted [22–24]. This issue may affect the fair evaluation of the algorithms.

Motivated by the issues mentioned above, in this paper, we first summarize and analyze the relat-ed metaheuristics studies. Then, extensive experiments are conducted by using representative benchmark functions, to fairly evaluate and understand the performances and characteristics of the state-of-the-art and the recent metaheuristics with a unified parameter tuning method. Furthermore, we test whether the algorithms have a search bias to the origin. Therefore, the main research contributions of our paper are outlined as follows:

- More than 500 metaheuristics are collected and a taxonomy of the metaheuristics is proposed. In particular, several representative algorithms are introduced from two aspects, including the inspiration sources and the essential operators for generating solutions.
- We perform extensive experiments to evaluate and understand the performances of the state-of-the-art and the recent metaheuristics. Eleven representative metaheuristics with new names (generally with high numbers of citations) and 4 state-of-the-art metaheuristics are selected to be comprehensively compared on the CEC2017 benchmark suite. In addition, whether these algorithms have a search bias to the origin is investigated. For fair comparisons, a unified framework named irace is used to tune the parameters of all the comparative algorithms.
- We use multiple nonparametric statistical methods to analyze the experimental results in depth. The statistical results show that the newly proposed EBCM algorithm performs similarly to the 4 compared algorithms and has the same properties and behaviors, such as convergence, diversity, exploration and exploitation trade-offs, in many aspects. However, the other 10 recent metaheuristics are less efficient and robust than the 4 state-of-the-art metaheuristics. All 15 algorithms show certain degrees of search bias toward the origin, but the 4 state-of-the-art metaheuristics are less affected by the shift operator on the functions. Furthermore, we find that the other 10 new algorithms (i.e., except for EBCM) are inferior to the 4 state-of-the-art algorithms in terms of convergence speed and global search ability on most of the CEC2017 functions. The other 10 new algorithms show a rougher trade-off and diversity compared with the 4 state-of-the-art algorithms. Finally, several important issues that should be considered in the metaheuristic research area are discussed and some potential research directions are suggested.

The paper is organized as follows: Section 2 presents a taxonomy of the metaheuristics and some representative metaheuristics are further introduced and investigated by explaining the inspiration sources and the essential operators for generating solutions. Extensive experiments are conducted to evaluate the performance of the 15 comparative algorithms in Section 3, and some properties of these algorithms are further studied, including convergence, diversity, and the exploration and exploitation trade-offs. Section 4 engages with some metaheuristics research issues and suggests several potential research directions. Section 5 draws the conclusion.

## 2 Literature overviews

In the last few decades, not only various improved versions of metaheuristics, but also many metaheuristics with new names mimicking the behaviors of humans, animals and plants, and the phenomena of physics and chemistry have been proposed. We selected some of the popular metaheuristics (i.e., 47 metaheuristics) to search for publications of these algorithms in the Web of Science updates to November 2021. Fig.1 shows that the number of publications for DE, PSO, SA, ACO, and the artificial immune system all exceeded 10000. It can be observed that many newly proposed metaheuristics have also received many citations and substantial attention. Therefore, metaheuristics are still among the hot research topics and it is expected that the number of publications of new metaheuristics and state-of-the-art metaheuristics will continue to increase in the future. Table 7[1] summarizes more than 500 metaheuristics, in which "B#" corresponds to reference [#] in Appendix B of the related supplemental material. For details on the full list of metaheuristics, please refer to the supplementary materials.

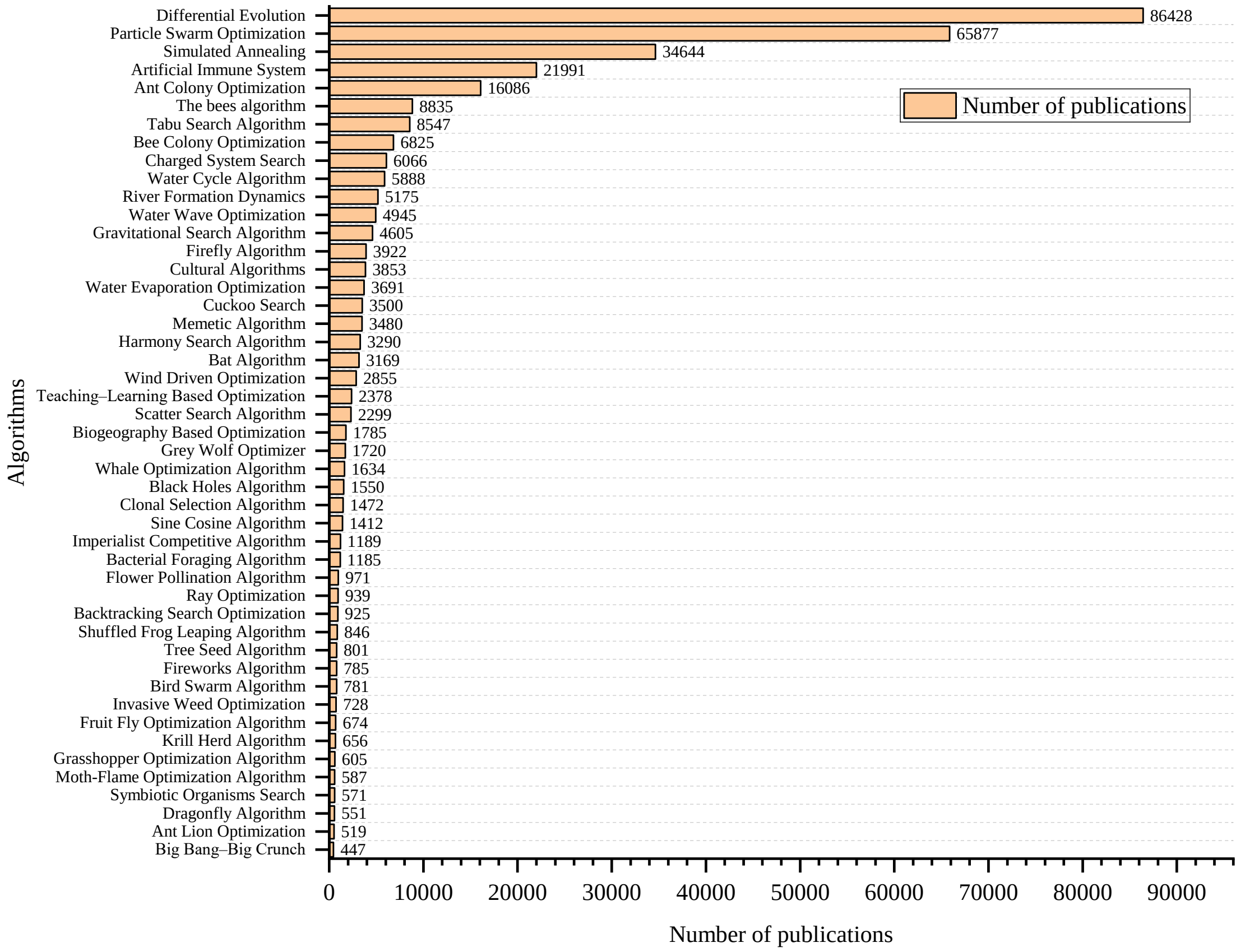

Fig.1 The number of publications about some popular metaheuristics

## 2.1 Taxonomy of metaheuristics

There are various taxonomy methods for metaheuristics in the literature, and the most popular taxonomy is based on the source of inspiration [19,20,25]. Fig.2 illustrates a rough metaheuristics classification, in which the metaheuristics are divided into population-based optimization algorithms (POAs) and single-solution based optimization algorithms (SOAs) according to the number of solutions generated in each iteration. SOAs generally require only one individual to search the solution space. In contrast, POAs contain multiple individuals that search the solution space cooperatively and globally with some operators and mechanisms, such as mutation, crossover, selection, information sharing, and search behavior learning.

We mainly focus on the POAs in this paper. Compared with SOAs, the most important characteristics of POAs are three-fold [1]. First, multiple points (i.e., solutions or individuals) are employed to search the solution space cooperatively. Second, mechanisms for information sharing and interactive learning among the individuals are adopted. Third, POAs are stochastic, as randomness is usually incorporated into search operators such as mutation and crossover. In Fig. 2, the POAs can be further roughly divided into evolutionary algorithms (EAs), swarm intelligence algorithms (SIAs) and physics or chemistry-based algorithms (P/CBAs).

---

[1] If a revision is required in Table 7, please inform the authors. Table 7 will be updated online at: https://github.com/P-N-Suganthan/MHA-500Plus

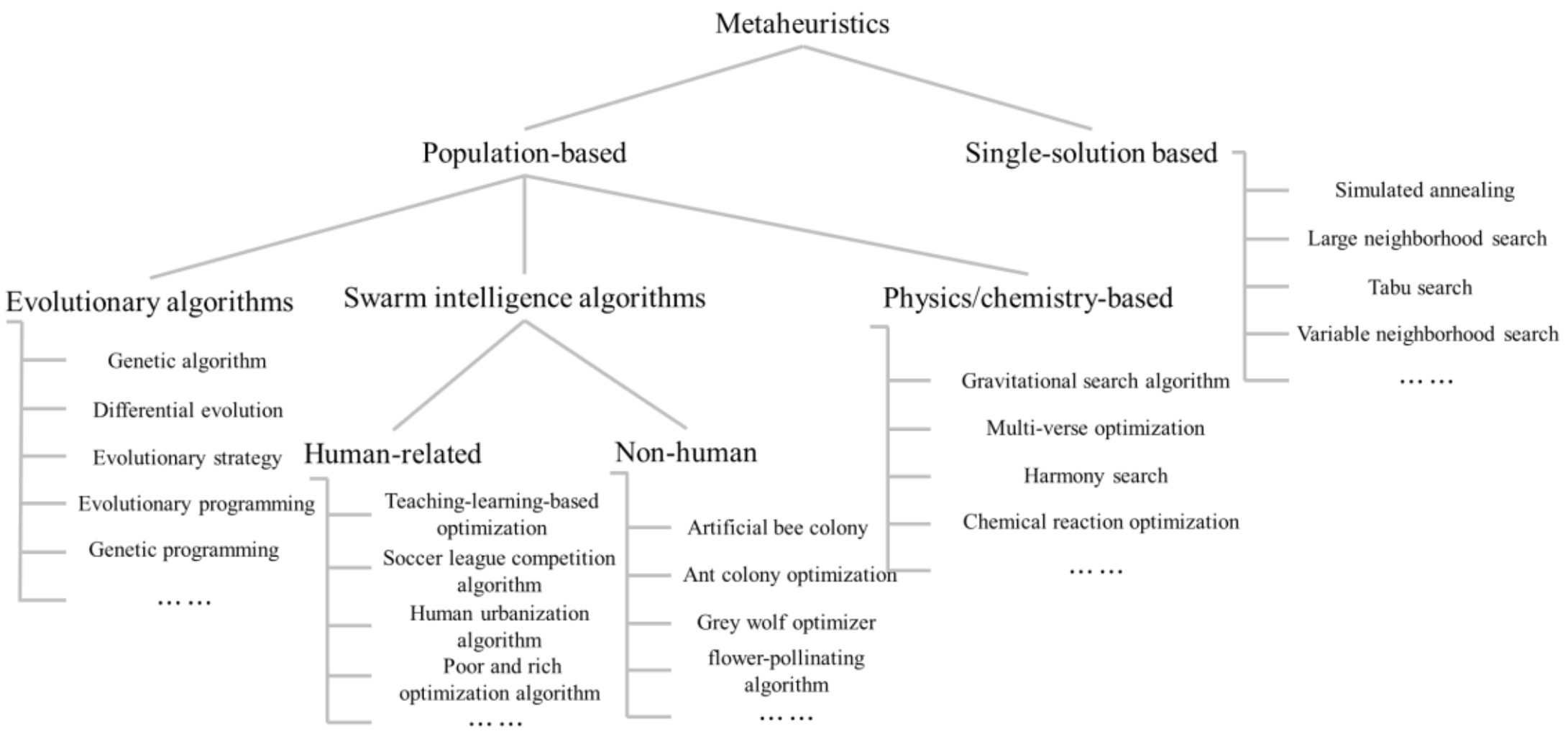

Fig.2 A classification of metaheuristics

### 2.1.1 Evolutionary algorithms

EAs are inspired by Darwinian evolutionary theory and mimic the behavior of evolution in nature, such as recombination, mutation, and selection [26], which fully embodies the idea of survival of the fittest. The first computer simulation of evolution can be traced back to 1954 by the work of Barricelli [27] but his publication did not attract widespread attention [28]. Until the 1960s and early 1970s, optimization methods could be designed via artificial simulated evolution after the use of evolutionary strategies (ES) to solve complex engineering problems in Rechenberg's work [29,30]. Currently, many variants of ES have been proposed in the literature, such as $(1+1)$-ES, $(\mu+1)$-ES, $(\mu+\lambda)$-ES, and $(\mu, \lambda)$-ES [31]. In 1960, evolutionary programming (EP) was first proposed by Fogel to achieve artificial intelligence [32,33]. Originally EP used finite state machines as predictors to predict environments. Currently, EP is a popular evolutionary algorithm and has many different versions including FEP (Fast EP) [34], AEP (Adaptive EP) [35], RLEP (Evolutionary Programming based on Reinforcement Learning) [36], and ENAEP (Ensemble algorithm of Gaussian and Cauchy mutation operators using AEP) [37]. In the early 1970s, genetic algorithms became popular through the work of Holland [15], and their performance mainly depended on the efficient encoding and decoding of the solution, appropriate parameter configuration and operators, including crossover, mutation and selection. GA and its variants are popular in a wide range of fields, such as planning [38] and scheduling [39], biological [40] and chemical [41] engineering, and data mining [42]. Later, genetic programming (GP) appeared and gradually became popular starting from the early 1990s. The variants of GP include MGP (multigene genetic programming) [43] , and GGP (grammatical genetic programming) [44]. Subsequently, differential evolution introduced by Storn and Price in 1995 [16,45], emerged as a very competitive evolutionary algorithm, especially in dealing with continuous optimization problems. There are many powerful and efficient variants of DE proposed in the literature, such as MPEDE [46] (multi-population ensemble DE), EDEV [47] (ensemble of multiple DE variants), SaDE [48] (with adapted mutation strategies and parameters), jDE [49] (with self-adapted parameters) and CoDE [50] (composition of multiple strategies and parameter settings).

### 2.1.2 Swarm intelligence algorithms

SIAs mimic the behaviors of animals, plants, and human groups in nature to optimize problems. Decentralized control and self-organization are two important features of SIAs [51], which can be understood as a group of individuals achieving common goals through cooperation. In other words, each individual of the swarm has its own intelligence and behaviors, and the integration of multiple individuals has more power to solve complex problems [52]. Particularly, the following advantages of SIAs make them user-friendly optimizers [53]: 1) The general framework can be applied to various fields with only a few modifications; 2) The information of the solution space and search states is reserved and used to guide the search during the optimization process; 3) Relatively fewer parameters make SIAs require less tuning effort to cater to different optimization problems. However, there still exist some critical issues that have not been well addressed in SIAs, such as premature convergence, being stuck in a local optimum, and lack of good trade-offs between exploitation and exploration [54]. Compared with EAs, SIAs do not have crossovers, while evolutionary algorithms usually have crossovers. SIAs do not include competitive selections, but EAs usually have selections. In addition, SIAs can be hybridized with EAs to include crossovers/selections. SIAs can be further categorized into human-related algorithms and nonhuman algorithms according to the inspiration source.

**(1) Human-related algorithms**

Human-related algorithms (HRAs) are inspired by the behaviors of humans in society, such as learning, competition, political campaigns, and cultural influence [55]. For example, inspired by the behavior of human learning, the teaching-learning-based optimization (TLBO) algorithm works on the effect of a teacher on learners [56]. The gaining-sharing knowledge-based algorithm (GSK) simulates the process of obtaining and sharing knowledge during the human lifespan [52]. The group teaching optimization algorithm (GTOA) mimics the mechanism of group teaching [57]. In terms of society competition, the soccer league competition algorithm (SLCA) is inspired by teams competing during a season in a soccer league [58] and the imperialist competitive algorithm (ICA) simulates the competition among imperialists [59]. Inspired by political campaigns, Askari et al. [5] conducted a comprehensive analysis of political mechanisms and proposed a new algorithm called the political optimizer (PO). In addition, the greedy reedy politics optimization (GPO) [60] and the parliamentary optimization algorithm (POA) [61] are also inspired by political mechanisms. There are many algorithms inspired by other human society inspiration sources, such as the poor and rich optimization algorithm (PRO) [62], human urbanization algorithm (HUS) [63], life choice-based optimizer (LCBO) [64] and queuing search algorithm (QS) [65].

**(2) Nonhuman algorithms**

Nonhuman algorithms (NHAs) include animal-based algorithms (AAs) and plant-based algorithms (PAs). AAs are inspired by the behavior of different animals, such as foraging, flocking, mating, and other behaviors [66]. For example, PSO is inspired by the behavior of a flock of birds or a school of fish, in which each particle can move throughout the solution space and update its current position in terms of a current best solution and a global best solution [53]. An artificial bee colony (ABC) is a metaheuristic based on the intelligent behavior of a honey bee swarm. The bee colony consists of three types of bees, employed bees, onlooker bees, and scout bees, and the search phases can be divided into search, recruit, and abandon [67]. The bat algorithm (BA) [68] and cuckoo search (CS) [69] are inspired by the echolocation behavior of bats and the brood parasitism of some cuckoo species, respectively. Other popular AAs include the firefly algorithm (FA) [70], gray wolf optimizer (GWO) [71] and grasshopper optimization algorithm (GOA) [72]. PAs are inspired by plant behavior such as growth, root expansion, weed invasion and flower pollinatio [65,73]. For instance, the invasive weed optimization (IWO) algorithm [74] mimics the process of weed invasion, and the flower-pollinating algorithm (FPA) [75] simulates the characteristics of flower pollination.

### 2.1.3 Physics/chemistry-based algorithms

P/CBAs are mostly created by imitating the physical and chemical law phenomena in nature, including electromagnetic force, inertia force, gravity, electrical charges, river systems, movement, chemical changes of material, and others [73,76–79]. For instance, the gravitational search algorithm (GSA) [80] is inspired by the law of gravity and mass interactions, where the search individuals are a collection of masses. According to the concepts of the white hole, black hole and wormhole in cosmology, multi-verse optimization (MVO) [81] has been designed to solve complex problems. In MVO, white holes and black holes are correlated with explorations, and wormholes are responsible for sharing and exploiting the information of the solution space. In addition, harmony search (HS) [82] mimics the behavior of an orchestra to create the most harmonious melody and measure it by aesthetic standards. Detailed information about HS is described in the literature [83]. Other typical P/CBAs include water evaporation optimization (WEO) [84], transient search optimization (TSO) [85], chemical reaction optimization (CRO) [86], and charged system search (CSS) [87].

## 2.2 Optimization mechanisms of metaheuristics

In this section, the optimization frameworks of single-solution based and population-based metaheuristics are presented. After that, several representative metaheuristics are reviewed from two different aspects, the inspiration sources and the essential operators for generating solutions.

### 2.2.1 Optimization framework of metaheuristics

As Algorithm 1 [88,89] shows, typical single-solution based metaheuristics start from a single initial solution. It iteratively performs a generation and selection procedure for a single solution until a termination condition is met; then, a best-so-far solution will be returned. In each iteration, a candidate solution set, $C(s_t)$, is generated based on the incumbent solution, $s_t$, in the generation procedure. In the selection phase, a selection operation is performed on the set $C(s_t)$ to choose a new solution $s_{t+1}$ to replace the current solution.

Population-based metaheuristics begin with an initial population solution $P_0$, as shown in Algorithm 2 [88,90]. Afterward, the generation and selection are iteratively executed to generate a new population $P_t'$, and selects promising individuals to form a new population $P_{t+1}$ to replace the current population. Finally, the best-so-far solution, $P^*$, is returned when a given stopping criterion is met. Moreover, the historical information can be memorized in Algorithm 1 and Algorithm 2 to better generate candidate solutions and to select promising solutions.

Regardless of the kinds of optimization frameworks and classification criteria used, exploration and exploitation play crucial roles in improving the performance of metaheuristics [47,91,92]. Exploration refers to the ability to globally search the solution space and find a promising region, which is associated with escaping from the local optimum and avoiding a premature convergence (i.e., increasing population diversity). Exploitation denotes the capability of locally searching the promising region found by the exploration operators. The well-known trade-off between exploration and exploitation is critical. In regards to the trade-offs of the exploration and exploitation of metaheuristics, Morales-Castañeda et al. [93] and Črepinšek et al.[94] conducted an in-depth investigation.

**Algorithm 1:** The common optimization framework of a single-solution based metaheuristics

Input: initial solution $s_0$; parameters

Output: The best solution

```
t ← 0;
Repeat
   /* Generate candidate solutions (partial or complete neighborhood) from s_t */
   Generate(C(s_t));
   /* Select a solution from C(s) to replace the current solution s_t */
```

| |
|---|
| $s_{t+1} = Select(C(s_t))$; |
| $t = t + 1$; |
| Until the termination condition is met. |

| **Algorithm 2:** The common optimization framework of a population-based metaheuristics |
|---|
| Input: initial solution $P_0$; parameters |
| Output: The best solution |
| t ← 0;<br>Evaluate the initial solutions and remember the best one as $P^*$;<br>Repeat<br>Generate $(P_t')$; /*Generation a new population */<br>$P_{t+1} = \text{Select_Population}(P_t \cup P_t')$; /*Select new population */<br>Record the best solution found so far $P^*$;<br>t ← t + 1;<br>Until the termination condition is met<br>return the best solution found $P^*$. |

### 2.2.2 Introduction of representative metaheuristics

In this section, several representative state-of-the-art and new metaheuristics are reviewed from two aspects: (1) the inspiration source for proposing the algorithm, and (2) the essential operators for generating solutions in each algorithm. These representative metaheuristics include some popular and competitive algorithms and the recently proposed algorithms.

(1) Differential Evolution (DE) [16]

DE is a competitive metaheuristic inspired by the principle of survival of the fittest. In DE, the population evolves through mutation, crossover, and selection in each generation, and the most frequently used mutation operator of DE is called DE/rand/1, which can be formulated as

$$\vec{v}_{i,G} = \vec{x}_{r_1^i,G} + F \cdot (\vec{x}_{r_2^i,G} - \vec{x}_{r_3^i,G}) \tag{1}$$

where $\vec{v}_{i,G}$ is the mutation vector, $\vec{x}_{r_1^i,G}$, $\vec{x}_{r_2^i,G}$, and $\vec{x}_{r_3^i,G}$ are three randomly generated distinct vectors, and $F$ is a mutation factor among [0,1].

The other popular mutation schemes are summarized as follows [95,96]:

$$\textit{DE/best/1}: \vec{v}_{i,G} = \vec{x}_{best,G} + F \cdot (\vec{x}_{r_1^i,G} - \vec{x}_{r_2^i,G}) \tag{2}$$

$$\textit{DE/best/2}: \vec{v}_{i,G} = \vec{x}_{best,G} + F \cdot (\vec{x}_{r_1^i,G} - \vec{x}_{r_2^i,G}) + F \cdot (\vec{x}_{r_3^i,G} - \vec{x}_{r_4^i,G}) \tag{3}$$

$$\textit{DE/rand/2}: \vec{v}_{i,G} = \vec{x}_{r_1^i,G} + F \cdot (\vec{x}_{r_2^i,G} - \vec{x}_{r_3^i,G}) + F \cdot (\vec{x}_{r_4^i,G} - \vec{x}_{r_5^i,G}) \tag{4}$$

$$\textit{DE/target-to-best/1}: \vec{v}_{i,G} = \vec{x}_{i,G} + F \cdot (\vec{x}_{best,G} - \vec{x}_{i,G}) + F \cdot (\vec{x}_{r_1^i,G} - \vec{x}_{r_2^i,G}) \tag{5}$$

$$\textit{DE/current-to-rand/1}: \vec{v}_{i,G} = \vec{x}_{i,G} + F \cdot (\vec{x}_{r_1^i,G} - \vec{x}_{i,G}) + F \cdot (\vec{x}_{r_2^i,G} - \vec{x}_{r_3^i,G}) \tag{6}$$

where $\vec{x}_{r_1^i,G}$, $\vec{x}_{r_2^i,G}$, $\vec{x}_{r_3^i,G}$, $\vec{x}_{r_4^i,G}$, and $\vec{x}_{r_5^i,G}$ are mutually different vectors, which are randomly chosen from the population at generation $G$. $\vec{x}_{i,G}$ is the target vector at generation $G$. $\vec{x}_{best,G}$ is the vector with the best fitness in the population at generation $G$. $F$ is the scaling factor within [0,1].

Two widely used crossover methods in DE are the binomial crossover and exponential crossover, and their formulas are shown as follows [97,98].

Binomial crossover:

$$u_{i,j,G} = \begin{cases} v_{i,j,G} & if\ rand_i(0,1) \le CR\ or\ j = j_{rand} \\ x_{i,j,G} & otherwise \end{cases} \tag{7}$$

Exponential crossover:

$$u_{i,j,G} = \begin{cases} v_{i,j,G} & for\ j = \langle l \rangle_D, \langle l+1 \rangle_D, \dots, \langle l+L-1 \rangle_D \\ x_{i,j,G} & otherwise \end{cases} \tag{8}$$

where $u_{i,j,G}$, $x_{i,j,G}$, and $v_{i,j,G}$ are the *j-th* components of vectors $\vec{u}_{i,G}$, $\vec{x}_{i,G}$, and $\vec{v}_{i,G}$, respectively, $i = \{1,2,\dots,NP\}$, and $j = \{1,2,\dots,D\}$. $j_{rand}$ is an integer, that is randomly generated in the range of $[1, D]$. $rand_i(0,1)$ is a number randomly generated from a uniform distribution in the range of [0,1]. The no-tation $<>_D$ denotes the modulo function with modulus $D$. $L$ is an integer number ranging in $[1, D]$.

The selection operation can be completed by comparing the fitness values of the target vector and the trial vector that determines which vectors can survive to the next generation.

$$\vec{x}_{i,G+1} = \begin{cases} \vec{u}_{i,G} & if\ f(\vec{u}_{i,G}) \leq f(\vec{x}_{i,G}) \\ \vec{x}_{i,G} & otherwise \end{cases} \tag{9}$$

where $\vec{x}_{i,G+1}$ and $\vec{x}_{i,G}$ are target vectors at the generations $G$ and $G+1$, respectively, $\vec{u}_{i,G+1}$ is the trial vector at generation $G$ and $f(*)$ is the objective function considered.

There exist many variants of DE in the literature. Some variants show competitive performance in solving complex problems. For example, LSHADE-cnEpSin [99] combines a sinusoidal approach based on performance adaptation and a covariance matrix learning method for the crossover operator into LSHADE-EpSin, which achieved a competitive performance at the 2017 IEEE CEC. Mohamed et al. [100] proposed a new version of DE named LSHADE-SPACMA by integrating LSHADE-SPA and a modified version of CMA-ES. Furthermore, Mohamed et al. [97] proposed two new DE variants named EDE and EBDE, where EDE had a less greedy mutation strategy named DE/current-to-ord_best/1, and EBDE introduced a more greedy mutation strategy named DE/current-to-ord_pbest/1. We named this newly proposed algorithm EDE-EBDE in our paper. MPEDE [46] and EDEV [47] are two powerful variants proposed by Wu. In MPEDE [46], three mutation strategies simultaneously co-existed, i.e., "current-to-pbest/1" and "current-to-rand/1" and "rand/1". EDEV [47] consists of three highly popular and efficient DE variants, namely JADE, CoDE, and EPSDE. The entire population of EDEV is partitioned into four subpopulations to coevolve to obtain better results. In the latest research, Ghosh et al. [101] combined SHADE and LSHADE with the nearest spatial neighborhood-based parameter adaptive process modification method to propose NSHADE and NLSHADE.

(2) Hybrid sampling evolution strategy (HSES) [102]

HSES is a new version of ES that combines the covariance matrix adaptation-evolution strategy (CMA-ES) and the univariate sampling method, it achieved the best performance at the 2018 IEEE CEC. In the HSES, the CMA-ES is mainly used for unimodal problems and the univariate sampling method is used for multimodal nonseparable problems. In particular, the method for calculating the mean value and the standard deviation of UMDAc (i.e., univariate marginal distribution algorithm continuous) is modified in the HSES. For detailed information about the HSES, please refer to [102].

(3) EBOwithCMAR [103]

EBOwithCMAR (Effective Butterfly Optimizer with Covariance Matrix Adapted Retreat Phase) is an improved variant of the butterfly optimizer, which combines a self-adaptive butterfly optimizer and a covariance matrix adapted retreat phase. The solution modification equations are shown as follows.

Criss-cross modification: $$\bar{v}_z = \bar{x}1_{cc_z} + F * (\bar{x}1_{r1_z} - (X_1 \cup X_2)_{r2_z}) \tag{10}$$

Toward-best modification: $$\bar{v}_z = \bar{x}1_{best_z} + F * (\bar{x}1_{cc_z} - (X_1 \cup X_2)_{r2_z}) \tag{11}$$

where $\bar{v}_z$ is a new vector, and $\bar{x}1_{cc_z}$, $\bar{x}1_{r1_z}$ and $(X_1 \cup X_2)_{r2_z}$ are three distinct individual vectors. $\bar{x}1_{best_z}$ is the best neighbor of the *z-th* vector. $F$ is a positive real number that controls the population evolution rate. $X_1 \cup X_2$ is the combination of both populations. In particular, the crossover operator of EBOwithCMAR is based on the equation (9).

(4) Snap-drift cuckoo search (SDCS) [104]

SDCS is a new version of CS [69] proposed by Rakhshani, that integrates the snap and drift modes into CS to establish the trade-off between exploration and exploitation. Moreover, a pair of new crossover and mutation operators are employed to improve the search capability. The updated rules of the SDCS are shown below.

Snap and drift modes
$$p_a = \begin{cases} \max(0, p_m - \omega) & if\ \mu = snap \\ \min(1, p_m + \omega) & if\ \mu = drift \end{cases} \quad (12)$$

Crossover operator
$$x_i^{t+1} = \begin{cases} x_i^t + a_0 \otimes (x_j^t \otimes Lévy(\beta) - x_i^t) & if\ p < J \\ x_i^t + a_0 \otimes (x_j^t - x_i^t \otimes Lévy(\beta)) & if\ J \leq p \leq 1 - J \\ x_i^t + a_0 \otimes (x_j^t - x_i^t) \otimes Lévy(\beta) & if\ p \geq 1 - J \end{cases} \quad (13)$$

Mutation operator
$$x_i^{t+1} = \begin{cases} x_i^t + H(p_a - \varepsilon) \otimes (x_j^t \otimes r - x_i^t) & if\ p < J \\ x_i^t + H(p_a - \varepsilon) \otimes (x_j^t - x_i^t \otimes r) & if\ J \leq p \leq 1 - J \\ x_i^t + H(p_a - \varepsilon) \otimes (x_j^t - x_i^t) & if\ p \geq 1 - J \end{cases} \quad (14)$$

where $p_a$ is known as a switching parameter [105], which is applied to trade-off the snap mode and drift mode. $p_m$ is the performance measure, and $\omega$ is the increase (or decrease) rate of $p_a$. $\mu$ is an auxiliary parameter that equals $snap$ if $0 \leq p_m \leq 0.5$; otherwise, $\mu = drift$. $x_i^t$ and $x_j^t$ are two mutually different positions at time $t$, and $x_i^{t+1}$ is the *i-th* position at time $t + 1$. $a_0$ and $\beta$ are the Lévy flight exponent and step size scaling factor, respectively. $H$ refers to the Heaviside step function. $r$, $p$, and $\varepsilon$ are three randomly generated numbers with uniform distributions, and $J \in [0,1]$ is a possibility value. The notation $\otimes$ indicates entrywise multiplications.

(5) Multi-strategy enhanced sine cosine algorithm (MSCA) [106]

MSCA is an improved version of SCA [107], which is based on sine and cosine functions and randomly generates multiple initial individuals to fluctuate outward or toward the best solution. In MSCA, multiple control mechanisms and operators are embedded into SCA, including the Cauchy mutation operator, chaotic local search mechanism, and opposition-based learning strategy, and two differential evolution operators are used to achieve a better trade-off between exploration and exploitation. The position update equations of MSCA are the same as those of SCA and can be expressed as follows.

$$X_i^{t+1} = \begin{cases} X_i^t + r_1 \times \sin(r_2) \times |r_3 P_i^t - X_i^t| & r_4 < 0.5 \\ X_i^t + r_1 \times \cos(r_2) \times |r_3 P_i^t - X_i^t| & r_4 \geq 0.5 \end{cases} \quad (15)$$

$$r_1 = a - t\frac{a}{T} \quad (16)$$

where $X_i^t$ is the position of the current solution in the *i-th* dimension at the *t-th* iteration, and $P_i^t$ is the position of the destination point in the *i-th* dimension at the *t-th* iteration. $r_1$ is a random variable that is calculated by Equation (16), which is a constant, $t$ is the current iteration, and $T$ is the maximum number of iterations. $r_2$ is a random variable responsible for the movement (i.e., toward or outward $P_i^t$) of the next solution (i.e., $X_i^{t+1}$). $r_3$ is a random variable that gives random weights for $P_i^t$. $r_4$ is a random number within [0,1]. For detailed information on the mechanisms and operators adopted in MSCA please refer to [106].

(6) Improved moth-flame optimization algorithm (IMFO) [108]

IMFO is a newly improved algorithm that introduces a hybrid phase, dynamic crossover mechanism, and fitness-dependent weight factor into MFO [109] to overcome the degeneration of the global search capability and convergence speed. The main inspiration of IMFO is also the navigation behavior of moths in nature, which is referred to as a transverse orientation. The position update equations of moths in IMFO are shown as follows.

$$w = \left| \frac{f(M_{best})}{f(M_i^k)} \right| \quad (17)$$

$$M_i^k = D_i^{k-1} e^{bt} \cos(2\pi t) + w \cdot F_i^{k-1} + (1 - w) \cdot M_{best} \quad (18)$$

where $w$ is a weight factor that depends on fitness, $f(M_{best})$ is the fitness value of the best solution $M_{best}$ and $f(M_i^k)$ represents the fitness values of the *i-th* moth at iteration $k$. $M_i^k$ and $F_i^{k-1}$ are the positions of the *i-th* moth and the *j-th* flame at iterations $k$ and $k - 1$, respectively. $D_i^{k-1}$ is the distance between the *i-th* moth and the *j-th* flame at iteration $k - 1$, $b$ is a constant used to define the shape of the logarithmic spiral and $t$ is a random number within [-1,1].

(7) Aquila optimizer (AO) [110]

AO is inspired by Aquila's behavior in nature during the process of catching prey. The optimization procedures of the proposed AO algorithm are represented in four methods, selecting the search space by a high soar with a vertical stoop, exploring within a divergent search space by a contour flight with a short glide attack, exploiting within a convergent search space by a low flight with a slow descent attack, and swooping by walking and grabbing the prey. In AO, the different steps (four methods) have different mathematical expressions for solution position updating and are shown as follows.

Step 1: $$X_1(t+1) = X_{best}(t) \times \left(1 - \frac{t}{T}\right) + (X_M(t) - X_{best}(t) * rand) \tag{19}$$

Step 2: $$X_2(t+1) = X_{best}(t) \times Levy(D) + X_R(t) + (y - x) * rand \tag{20}$$

Step 3: $$X_3(t+1) = (X_{best}(t) - X_M(t)) \times \alpha - rand + (UB - LB) \times rand + LB) \times \delta \tag{21}$$

Step 4: $$X_4(t+1) = QF \times X_{best}(t) - (G_1 \times X(t) \times rand) - G_2 \times Levy(D) + rand \times G_1 \tag{22}$$

where $X_i(t+1)$ is the solution of the next iteration of $t$, which is generated by each search method $(X_I)$ and $i = 1,2,3,4$. $X_{best}(t)$ is the best-obtained solution until the $t^{th}$ iteration, $X_M(t)$ denotes the location mean value of the current solutions at the $t^{th}$ iteration, $Levy(D)$ is the levy flight distribution function, $X_R(t)$ is a random solution taken in the range of $[1, N]$ at the $t^{th}$ iteration, $y = r \times \cos(\theta)$ and $x = r \times \sin(\theta)$ are used to present the spiral shape in the search and $\alpha$ and $\delta$ are the exploitation adjustment parameters fixed at 0.1. $LB$ and $UB$ are the lower bound and upper bound of the given problem, respectively. $QF$ denotes a quality function used to establish the equilibrium of the search strategies, $G_1$ denotes various motions of the AO, $G_2$ denotes the flight slope of the AO that is used to follow the prey during the slope from the first location (1) to the last location $(t)$, $rand$ is a random value between 0 and 1, and $t$ and $T$ represent the current iteration and the maximum number of iterations, respectively. For detailed parameter information calculations and the AO, please refer to [110].

(8) Improved grasshopper optimization algorithm (IGOA) [111]

The grasshopper optimization algorithm (GOA) [71] is a recently proposed metaheuristic algorithm that is inspired by the swarming behavior of grasshoppers. IGOA improves GOA through the integration of multiple mechanisms including Gaussian mutation, Levy-flight strategy and opposition-based learning. The improvement method in IGOA is similar to that of the MSCA. The mathematical expression of the solution position update is as follows.

$$X_i^d = c\left(\sum_{\substack{j=1 \\ j \neq i}}^{N} c\frac{ub_d - lb_d}{2} s(|x_j^d - x_i^d|)\frac{x_j - x_i}{d_{ij}}\right) \oplus G(\alpha) + \hat{T}_d \tag{23}$$

$$X_i^{levy} = X_i^* + rand(d) \oplus levy(\beta) \tag{24}$$

$$X_i^{t+1} = \begin{cases} X_i^{levy} & fitness(X_i^{levy}) > fitness(X_i^*) \\ X_i^* & otherwise \end{cases} \tag{25}$$

In Equation (23), $X_i^d$ represents the updated position of grasshopper $i$ in the *d-th* dimension. $x_j^d$ and $x_i^d$ are two different grasshoppers in the *d-th* dimension. $N$ is the number of grasshoppers, and $c$ is a parameter calculated by the equation $c = cmax - l\,(cmax - cmin)/L$. $ub_d$ and $lb_d$ are the upper bound and lower bound in the *d-th* dimension, respectively. $s(*)$ is the function that defines the social forces, and $d_{ij}$ is the distance between the *i-th* and *j-th* grasshoppers. $G(*)$ and $\hat{T}_d$ are the Gaussian step vector and the value of the best-so-far solution in the *d-th* dimension, respectively. $\alpha$ is a Gaussian random number generated in the range of [0,1]. The generation and selection of the new solution are based on Equations (24) and (25). $X_i^{levy}$ is a new solution generated based on the Levy flight mechanism, and $X_i^*$ is the new position of the *i-th* grasshopper after updating. $X_i^{t+1}$ is a selected solution based on the fitness value between $X_i^{levy}$ and $X_i^*$. $rand(d)$ and $levy(*)$ are the *d-th* dimension random vectors in [0,1] and the Levy distribution, respectively. $\beta$ is the Levy index. The notation $\oplus$ in all the equations represents the dot product operation.

(9) Hyperbolic gravitational search algorithm (HGSA) [112]

GSA is a physically inspired population-based algorithm that solves problems based on the law of gravity and mass interactions [80]. HGSA is a new version of GSA, in which the hyperbolic acceleration coefficient, dynamic regulation, and decreasing hyperbolic function are adopted to achieve a better trade-off between exploration and exploitation. The positions and velocities of the individuals in HGSA can be calculated as follows.

$$v_i^d(t+1) = rand_i \times v_i^d(t) + c_1(t) \times a_i^d(t)\Delta t + c_2(t) \times (gbest - x_i^d(t))/\Delta t \tag{26}$$

$$x_i^d(t+1) = x_i^d(t) + v_i^d(t+1) \tag{27}$$

where $x_i^d(t)$ and $v_i^d(t)$ are the position and velocity of the *i-th* individual in the *d-th* dimension at iteration $t$, $a_i^d(t)$ is the acceleration of individual $i$ at time $t$ and $c_1(t)$ and $c_2(t)$ are the acceleration coefficients at time $t$. $rand_i$ is a uniform random variable in the interval [0,1]. $gbest$ is the position of the best-so-far solution. $\Delta t$ is the time increment. For detailed information about HGSA, please refer to [112].

(10) Memetic frog leaping algorithm (MFLA)

MFLA is an improved version of the shuffled frog leaping algorithm (SFLA) that was first proposed by Eusuff et al. [113]. SFLA is a metaheuristic search approach that mimics the foraging behavior of frogs, which is similar to PSO. In the frog population, each frog can communicate with each other and the worst frog can jump to find the best food source guided by the best frog. MFLA improves SFLA by integrating a memetic mechanism and a new search leaping rule. The mathematical formulas are shown below.

$$Q_m = \begin{cases} Q_g & if\ rand < 0.5 \\ Q_C & else \end{cases} \tag{28}$$

$$\acute{Q_w} = Q_w + rand(Q_{best} - Q_w) + rand(Q_m - Q_w) \tag{29}$$

where $Q_w$ and $Q_{best}$ are the worst and best frog, respectively. $Q_m$ is an auxiliary variable. $Q_g$ and $Q_C$ represent the geometric center and gravitational center, respectively. For the equations for calculating $Q_g$ and $Q_C$, please refer to [114].

(11) Gaining-sharing knowledge-based algorithm (GSK) [52]

GSK is inspired by the human behaviors of gaining and sharing knowledge, which can be divided into two phases: (1) the junior gaining and sharing phase and (2) the senior gaining and sharing phase. The differential equations for generating the new solutions are proposed in two phases and described as follows:

$$x_{ij}^{new} = \begin{cases} x_i + k_f * [(x_{i-1} - x_{i+1}) + (x_r - x_i)] & f(x_i) > f(x_r) \\ x_i + k_f * [(x_{i-1} - x_{i+1}) + (x_i - x_r)] & f(x_i) \le f(x_r) \end{cases} \tag{30}$$

$$x_{ij}^{new} = \begin{cases} x_i + k_f * [(x_{p-best} - x_{p-worst}) + (x_m - x_i)] & f(x_i) > f(x_m) \\ x_i + k_f * [(x_{p-best} - x_{p-worst}) + (x_i - x_r)] & f(x_i) \le f(x_m) \end{cases} \tag{31}$$

where $x_i$ is the *i-th* individual. Compared with the current individual $x_i$, $x_{i-1}$ and $x_{i+1}$ are the nearest better and worse individuals respectively, to constitute the gain source of knowledge. $x_{p-best}$ and $x_{p-worst}$ are the best individual and worst individual, respectively, among all the individuals. $x_r$ and $x_m$ are individuals randomly selected from the population, $k_f$ is a real number greater than 0 and $f(*)$ is the objective function.

(12) Marine predators algorithm (MPA) [115]

MPA mimics marine predators and uses the predation behavior of the Lévy and Brownian movements to optimize problems. The optimal encounter rate policy in the interaction between predator and prey is also considered. In MPA, the optimization process is divided into three phases due to different velocity ratios.

Phase 1: When the velocity ratio is high or the prey is moving faster than the predator

$$\overrightarrow{stepsize_i} = \vec{R}_B \otimes \left(\overrightarrow{Elite_i} - \vec{R}_B \otimes \overrightarrow{Prey_i}\right) i = 1, \cdots n \tag{32}$$

$$\overrightarrow{Prey_i} = \overrightarrow{Prey_i} + P \cdot \vec{R} \otimes \overrightarrow{stepsize_i} \tag{33}$$

Phase 2: In the unit velocity ratio or when both predator and prey are moving at almost the same pace

- For the first half of the population ($i = 1, \cdots n / 2$)

$$\overrightarrow{stepsize_i} = \vec{R}_L \otimes \left(\overrightarrow{Elite_i} - \vec{R}_L \otimes \overrightarrow{Prey_i}\right) i = 1, \cdots n / 2 \tag{34}$$

$$\overrightarrow{Prey_i} = \overrightarrow{Prey_i} + P \cdot \vec{R} \otimes \overrightarrow{stepsize_i} \tag{35}$$

- For the second half of the population ($i = n / 2, \cdots n$)

$$\overrightarrow{stepsize_i} = \vec{R}_B \otimes \left(\vec{R}_B \otimes \overrightarrow{Elite_i} - \overrightarrow{Prey_i}\right) i = n / 2, \cdots n \tag{36}$$

$$\overrightarrow{Prey_i} = \overrightarrow{Elite_i} + P \cdot CF \otimes \overrightarrow{stepsize_i} \tag{37}$$

Phase 3: In a low-velocity ratio when the predator is moving faster than the prey

$$\overrightarrow{stepsize_i} = \vec{R}_L \otimes \left(\vec{R}_L \otimes \overrightarrow{Elite_i} - \overrightarrow{Prey_i}\right) i = 1, \cdots n \tag{38}$$

$$\overrightarrow{Prey_i} = \overrightarrow{Elite_i} + P \cdot CF \otimes \overrightarrow{stepsize_i} \tag{39}$$

where $\overrightarrow{stepsize_i}$ is the step size matrix of the search individuals (predator and prey) and $\overrightarrow{Elite_i}$ is the matrix that oversees the searching and finding of the prey based on the information of the prey's positions. $\overrightarrow{Prey_i}$ is the function matrix based on which the predators update their positions. $\vec{R}_B$ is a vector containing random numbers that represents a Brownian motion. $P$ is the constant number, and $\vec{R}$ is a vector that contains a random number in the interval [0,1]. $\vec{R}_L$ is a vector that contains random numbers following the Lévy distribution. $CF$ is an adaptive parameter used to control the predator step size.

(13) Equilibrium optimizer (EO) [91]

EO is inspired by the control volume mass balance models that are used to estimate both the dynamic and equilibrium states. In EO, each individual has its concentration (position), and the best-so-far solution is named the equilibrium candidate. Each individual randomly updates their concentration around the equilibrium candidates to finally reach the equilibrium state (optimal result). The updating rule of the individuals is shown as follows.

$$\vec{C} = \vec{C}_{eq} + \left(\vec{C} - \vec{C}_{eq}\right) \cdot \vec{F} + \frac{\vec{G}}{\vec{\lambda} V}(1 - \vec{F}) \tag{40}$$

where $\vec{C}$ is a concentration vector of the individuals, $\vec{C}_{eq}$ is a vector that contains the candidates in the equilibrium pool, $\vec{F}$ is an exponential term that includes an exponential function, $V$ is considered a unit, and $\vec{\lambda}$ is a random vector in the range of [0,1]. In EO, the selection process is completed by comparing the fitness value of $\vec{C}_i$ with the fitness values of $\vec{C}_{eq1}$, $\vec{C}_{eq2}$, $\vec{C}_{eq3}$, and $\vec{C}_{eq4}$ and selecting the best one to replace the worst one.

As mentioned above, we can summarize the general mathematical model of the essential operators that generate solutions in the population-based metaheuristics, which can be described as the new solution $x_i^{t+1}$ being equal to the sum of the current solution $x_i^t$ and the modification increment or mutation vector $\Delta x_i^t$ [78], i.e.,

$$x_i^{t+1} = x_i^t + \Delta x_i^t \tag{41}$$

The ways to determine $\Delta x_i^t$ reflect the essential differences among the different metaheuristics.

# 3 Experimental analyses

To evaluate the performance and properties of the newly proposed algorithm, 11 newly named metaheuristics and 4 state-of-the-art metaheuristics are selected in this section. We first use a unified framework named irace to automatically configure the parameters of all 15 comparative algorithms.

Then, whether these algorithms have a search bias to the origin is investigated. For detailed description, the convergence, diversity, and trade-off between the exploration and exploitation of all 15 algorithms are also analyzed. All the experimental results were analyzed by nonparametric statistical methods, including the Friedman test, Wilcoxon signed-ranks test, and Bayesian signed-rank test.

## 3.1 Experiment setup

In this section, 11 newly proposed representative metaheuristics that are popular and highly cited and 4 state-of-the-art metaheuristics are selected for the comparison experiments. The 15 algorithms are summarized in Table 1.

Table 1 Summary of the 15 comparison algorithms

| Algorithm | Year | Abbreviation |
|---|---|---|
| ● 11 new algorithms | | |
| Aquila optimizer [110] | 2017 | AO |
| Effective butterfly optimizer with covariance matrix adapted retreat phase [103] | 2017 | EBOwithCMAR (EBCM) |
| Snap-drip cuckoo search [104] | 2017 | SDCS |
| Improved grasshopper optimization algorithm [111] | 2018 | IGOA |
| Hyperbolic gravitational search algorithms [112] | 2019 | HGSA |
| Memetic frog leaping algorithm [114] | 2019 | MFLA |
| Improved moth-flame optimization algorithm [108] | 2020 | IMFO |
| Multi-strategy enhanced Sine Cosine Algorithm [106] | 2020 | MSCA |
| Gaining-sharing knowledge-based algorithm [52] | 2019 | GSK |
| Marine predators algorithm [115] | 2020 | MPA |
| Equilibrium optimizer [91] | 2020 | EO |
| ● 4 state-of-the-art algorithms | | |
| L-SHADE with nearest spatial neighborhood-based modification [101] | 2017 | NLSHADE |
| LSHADE with semi-parameter adaptation hybrid with CMA-ES [100] | 2017 | LSHADE-SPACMA (LS-SPA) |
| Hybrid sampling evolution strategy [102] | 2018 | HSES |
| Two enhanced DE variants EDE and EBDE [97] | 2019 | EDE-EBDE (ED-EB) |

Note: In the rest of this paper, we use EBCM, LS-SPA, and ED-EB to represent EBOwithCMAR, LSHADE-SPACMA, and EDE-EBDE, respectively.

We select 2017 CEC bound-constrained numerical optimization problems as the benchmark problems[116], which contain thirty functions that can be divided into four categories, unimodal functions (F1-F3), multimodal functions(F4-F10), hybrid functions (F11-F20) and composition functions (F21-F30). These functions have the same upper bound (100) and lower bound (-100). The global minimum value of each function is the product of the function index and 100. In this paper, the maximum number of function evaluations is set to $10000 * D$, and all the experimental results are obtained from average values over 31 runs.

To make fair comparisons [117,118], we first tuned the parameters of all 15 comparative algorithms on all the CEC 2017 functions with 10, 30, and 50 variables. Afterward, further experiments are conducted from two aspects: performance evaluation and verification of whether these comparison algorithms have a search bias to the origin. In the performance evaluation experiments, all the CEC 2017 functions with 10, 30, and 50 variables are used. To evaluate whether these comparative algorithms have a search bias to the origin, all the shifted and nonshifted CEC 2017 functions with 10 and 30 variables are used.

All experimental results are analyzed by several nonparametric statistical methods including the Bayesian rank-sum test, Friedman test, Wilcoxon signed-rank test, and Bayesian signed-rank test to verify whether the performance of two or more algorithms differs from each other statistically. If the $p$ value obtained by any two algorithms is equal to or less than 0.05, it indicates that there is a significant difference in the performance of the two algorithms. Otherwise, the opposite is true. Details of the statistical results are summarized in the supplementary file, where the result tables (or figures) are denoted as “TableS# (Fig.S#)” and “#” is the table number. All the algorithms are coded in MATLAB software and run on a Windows 10 operating system with a Core i7-10700CPU and 32G RAM. The codes of this paper have been published online (http://faculty.csu.edu.cn/guohuawu/zh_CN/zdylm/193832/list/index.htm) .

## 3.2 Automatic parameter tuning

For a fair comparison, we employed the iterated racing (irace) method to automatically configure the main parameters. The iterated racing method repeats three steps until it meets a termination criterion: (1) sampling new configurations according to a particular distribution; (2) selecting the best configurations from the newly sampled configurations by means of racing, and (3) updating the sampling distribution to bias the sampling toward the best configurations [119].

The irace method is implemented through an R package named irace, developed by López-Ibáñez et al. [119]. Irace receives as input a parameter space definition corresponding to the parameters of our 15 algorithms that will be tuned, a set of training instances for which the parameters must be tuned, and a set of options for the irace that define the configuration scenario. For example, we tune the AO parameters, and the training instances are eight functions covering all types of the CEC 2017 functions. Then, the irace searches in the parameter search space for good-performing algorithm configurations by executing AO on different functions with different parameter configurations. In other words, all the parameter configurations will be tested on all the functions to verify which is the best performing configuration. For a detailed implementation of the irace method, please refer to [119–123].

## Table 2 Parameter adjustment results of the 15 algorithms

| Algorithm | Default parameters | Tuned parameters | | |
|---|---|---|---|---|
| | | 10 variables | 30 variables | 50 variables |
| AO [110] | Population size $n$=25; number of clusters $m$= 5; | Population size $n$=34; Exploitation adjustment parameters $\alpha$=0.9161; $\delta$=0.3806 | Population size $n$=10; Exploitation adjustment parameters α=0.4207; δ=09379 | Population size $n$=69; Exploitation adjustment parameters α=0.186; δ=0.6773 |
| SDCS [104] | Population size $n$={15, 25, 35}; Increase/decrease rate of $p_a$={0.005, 0.5, 1}; Movement variability parameter $J$={0.1, 0.2, 0.3}; Step size scaling factor $a_0$={0.01, 0.1, 1} | Population size $n$=10; Increase/decrease rate of $p_a$ =0.3413; Movement variability parameter $J$=0.8281; Step size scaling factor $a_0$=0.9491 | Population size $n$=24; Increase/decrease rate of $p_a$ =0.1854; Movement variability parameter $J$=0.9618; Step size scaling factor $a_0$=0.5973 | Population size $n$=10; Increase/decrease rate of $p_a$ =0.9137; Movement variability parameter $J$=0.9316; Step size scaling factor $a_0$=0.5201 |
| IGOA [111] | Population size $n$ =30 | Population size $n$=34 | Population size $n$=35 | Population size $n$=25 |
| HGSA [112] | Population size $n$=30; Gravitational constant $G_0$=100 | Population size $n$=37; Gravitational constant $G_0$=89 | Population size $n$=23; Gravitational constant $G_0$=118 | Population size $n$=24; Gravitational constant $G_0$=116 |
| MFLA [114] | Number of memeplexes $m$={2, 4, 5}; Number of frogs in a memeplex $n$={4, 5, 10}; *beta*=0.6 | Number of memeplexes $m$=5; Number of frogs in a memeplex $n$=5; *beta*=0.7563 | Number of memeplexes $m$=4; Number of frogs in a memeplex $n$=6; *beta*=0.5867 | Number of memeplexes $m$=4; Number of frogs in a memeplex $n$=5; *beta*=1.4742 |
| IMFO [108] | Population size $n$=100; Spiral shape parameter $b$=1; Iteration ratio $P$=0.5 | Population size $n$=119; Spiral shape parameter $b$=4; Iteration ratio $P$=0.0199 | Population size $n$=118; Spiral shape parameter $b$=4; Iteration ratio $P$=0.2963 | Population size $n$=93; Spiral shape parameter $b$=3; Iteration ratio $P$=0.3593 |
| MSCA [106] | Population size $n$ =30; Probability factor $P_c$=0.8; Constant number $a$=2; $\mu$=4 is a parameter that controls the degree of chaotic function. | Population size $n$=27; Probability factor $P_c$=0.0659; Constant number $a$=1; $\mu$=3 is a parameter that controls the degree of chaotic function. | Population size $n$=31; Probability factor $P_c$=0.0319; Constant number $a$=1; $\mu$=4 is a parameter that controls the degree of chaotic function. | Population size $n$=31; Probability factor $P_c$=0.0116; Constant number $a$=1; $\mu$=4 is a parameter that controls the degree of chaotic function. |
| GSK [52] | Population size $n$=100; Top and bottom percentage of individuals $P$=0.1; Knowledge factor $k_f$ =0.5; Knowledge ratio $k_r$=0.9; Knowledge rate $K$=10 | Population size $n$ =101; Top and bottom percentage of individuals $P$=0.1353; Knowledge factor $k_f$ =0.4822; Knowledge ratio $k_r$ =0.9797; Knowledge rate $K$=12 | Population size $n$ =93; Top and bottom percentage of individuals $P$=0.052; Knowledge factor $k_f$=0.485; Knowledge ratio $k_r$=0.991; Knowledge rate $K$=10 | Population size $n$ =100; Top and bottom percentage of individuals $P$=0.0521; Knowledge factor $k_f$ =0.4581; Knowledge ratio $k_r$ =0.9309; Knowledge rate $K$=9 |
| MPA [115] | Population size $n$=25; Probability factor *FADs*={0.1, 0.2, 0.5, 0.7, 0.9}; Constant number $P$={0.1, 0.5, 1, 1.5, 2} | Population size $n$=21; Probability factor *FADs*=0.8297; Constant number $P$=0.6737 | Population size $n$=31; Probability factor *FADs*=0.1014; Constant number $P$=0.1949 | Population size $n$=25; Probability factor *FADs*=0.3425; Constant number $P$=0.5076 |
| EO [91] | Population size $n$=30; Constant number $a_1$ =2; Constant number $a_2$ =1; Generation Probability $GP$=0.5 | Population size $n$=33; Constant number $a_1$=1.8876; Constant number $a_2$=0.9305; Generation Probability $GP$=0.2999 | Population size $n$=31; Constant number $a_1$=1.9447; Constant number $a_2$=0.95021; Generation Probability $GP$=0.5871 | Population size $n$=20; Constant number $a_1$=1.8587; Constant number $a_2$=1.1681; Generation Probability $GP$=0.7087 |
| EBCM [103] | $prob_{ls}$ =0.1; $\sigma$ =0.3; *arch_rate*=2.6; Memory size $H$=6 | $prob_{ls}$ =0.9209; $\sigma$ =0.2997; *arch_rate*=2.3947; Memory size $H$=5 | $prob_{ls}$ =0.4149; $\sigma$ =0.9267; *arch_rate*=3.2152; Memory size $H$=8 | $prob_{ls}$ =0.818; $\sigma$ =0.019; *arch_rate*=3.0527; Memory size $H$=4 |
| NLSHADE [101] | Population size $N_p$ = {50,100}; $M_F$ and $M_{Cr}$ are memory archive values, $M_F$ = 0.5, $M_{Cr}$ = 0.5. | Population size $N_p$= 140; $M_F$ and $M_{Cr}$ are memory archive values, $M_F$ = 0.8404, $M_{Cr}$ = 0.9969. | Population size $N_p$= 138; $M_F$ and $M_{Cr}$ are memory archive values, $M_F$ = 0.897, $M_{Cr}$ = 0.7163. | Population size $N_p$= 164; $M_F$ and $M_{Cr}$ are memory archive values, $M_F$ = 0.9039, $M_{Cr}$ = 0.792. |
| HSES [100] | Population size $M$=200; $N$=100 | Population size $M$=182; $N$=90 | Population size $M$=181; $N$=98 | Population size $M$=214; $N$=92 |
| LS-SPA [102] | $pbest$ =0.3; $pbest_{min}$=0.15 | $pbest$ =0.416; $pbest_{min}$=0.1732 | $pbest$ =0.4765; $pbest_{min}$=0.1459 | $pbest$ =0.2438; $pbest_{min}$=0.1749 |
| ED-EB [97] | *L_Rate*=0.8; *EDE_best_rate*=0.1; *Memory size*=5 | *L_Rate*=0.;0.2797 *EDE_best_rate*=0.4957; *Memory_size*=5 | *L_Rate*=0.7763; *EDE_best_rate*=0.1264; *Memory_size*=6 | *L_Rate*=0.5034; *EDE_best_rate*=0.2124; *Memory_size*=4 |

Note: The adjustment parameters of each algorithm are determined based on the published paper and the codes are obtained from the authors' websites.

## 3.3 Experimental Results and Discussion

In this section, 11 recent algorithms and 4 state-of-the-art algorithms are compared on CEC 2017 functions with 10, 30, and 50 variables, respectively. The experimental results of the 15 algorithms are summarized in Tables S1-S3. Nonparametric statistical methods, including Friedman test, Bayesian signed-rank, and Wilcoxon signed-rank test. The detailed statistical results can be found in Tables S4-S22 of the supplemental materials. Due to space limitations, we only show the analysis results on functions with 30 variables in the text. For more information about the analysis results on functions with 10 and 50 variables, please refer to Sections 3 and 4 of Appendix A in the supplementary material.

### 3.3.1 Benchmark functions with 30 variables

#### 3.3.1.1 Comparison of each function

Some interesting observations can be obtained from the statistical results of the functions with 30 variables reported in Table S2. It is observed that MFLA, GSK, IMFO, MPA, AO, and EBCM exhibit competitive performance among the 11 recent algorithms as compared with the 4 state-of-the-art algorithms. In particular, EBCM has the best performance among the 11 newly proposed algorithms. EBCM outperforms HSES, ED-EB, LS-SPA and NLSHADE on twenty-one (F2, F5-F6, F7-F9, F11, F13, F15, F16-F17, F20-F21 and F23-F30), fourteen (F2, F4, F5, F7-F8, F11-F13, F16-F17, F21, F25-F26, and F28), fourteen (F5, F7-F8, F10-F11, F13, F16, F21, F23-F26, F28 and F30), and five (F5-F6, F8, F13 and F25) functions, respectively.

MFLA is superior to HSES, ED-EB, LS-SPA and NLSHADE on eight (F1-F2, F6, F20, and F25-F28), four (F4 and F25-F26), four (F4 and F25-F26), and two (F22 and F25) functions. Moreover, MFLA exhibits high efficiency in dealing with the composition function F25.

GSK is superior to HSES, ED-EB, LS-SPA and NLSHADE on fifteen (F1-F4, F6, F11, F13, F15-F16, F20, F23, and F26-F29), three (F4, F11, and F28), five (F4, F23-F24, F26, and F28), and two (F22 and F25) functions. Particularly, the performance of GSK is equivalent to the 4 state-of-the-art algorithms (except NLSHADE) in solving the composition function F22. Moreover, GSK is superior or similar to all 5 competition algorithms on multimodal function F4 and composition function F28.

IMFO outperforms HSES, ED-EB, LS-SPA, and NLSHADE on five (F1-F2, F6, F20, and F25), two (F4 and F25), two (F4 and F25), and one (F25) functions.

The MPA achieves better results than the HSES, ED-EB, LS-SPA, and NLSHADE on seven (F2, F6, F20-F21, F23, F26, and F28), three (F4, F26, and F28), five (F4, F21, F23, F26, and F28), and one (F25) functions. Particularly, MPA yields promising performance on the composition function F26 by surpassing all 4 state-of-the-art algorithms.

In contrast, EO, AO, HGSA, IGOA, MSCA, and SDCS demonstrate less efficiency than the 4 state-of-the-art algorithms. For instance, these 6 recent algorithms are only superior to the 4 state-of-the-art algorithms in less than 3 functions. In particular, AO is almost inferior to the 4 state-of-the-art algorithms on all thirty functions.

In conclusion, EBCM shows competitive performance compared with 4 state-of-the-art algorithms. The performance of EBCM completely surpasses HSES and is comparable to ED-EB and LS-SPA on the CEC 2017 functions with 30 variables. However, MFLA, GSK, IMFO, and MPA are inferior to the 4 state-of-the-art algorithms in most functions. EO, AO, HGSA, IGOA, MSCA, and SDCS have less efficiency in dealing with CEC 2017 functions with 30 variables since they are only superior to or comparable to the 4 state-of-the-art algorithms in a few functions. The results show that MFLA/GSK/IMFO/MPA/EBCM is superior to HSES, ED-EB, LS-SPA, and NLSHADE on 8/4/4/2/21, 15/3/5/2/14, 5/2/2/1/14, and 7/3/3/1/5 functions, respectively. In addition, MFLA and GSK are compa-rable to these 3 comparative algorithms (except NLSHADE) on function F22. EO, AO, HGSA, IGOA, MSCA, and SDCS only perform better than HSES, ED-EB, LS-SPA, and NLSHADE on 2/1/1/0, 0/0/0/1, 1/1/2/1, 0/0/0/1, 2/0/1/1, and 0/1/1/1 function(s). It is worth noting that 11 recent algorithms become less efficient as the dimension of the functions increases (i.e., from 10 variables to 30 variables).

#### 3.3.1.2 Results of Wilcoxon signed-rank test

As seen from Table 3, EBCM performs competitively with 4 state-of-the-art algorithms on the CEC 2017 functions with 30 variables. In addition, GSK exhibits significantly similar performance to HSES in solving functions with 30 variables which is consistent with the conclusion of the Bayesian rank-sum test and the Friedman test. In contrast, the performances of EO, AO, HGSA, IGOA, IMFO, MFLA, MPA, MSCA, SDCS, and HSES are significantly different from those of the 4 state-of-the-art algorithms. In other words, these 9 recent algorithms are not efficient in dealing with the CEC 2017 functions with 30 variables. It is worth noting that some recent algorithms, such as MPA, SDCS and MFLA, demonstrate high efficiency on functions with 10 variables but have a deteriorated performance in solving functions with 30 variables.

Table 3 The results with significant differences of the Wilcoxon signed-rank test for 30 variables

| Algorithms | 30 variables | | |
|---|---|---|---|
| | $R^+$ | $R^-$ | *p-value* |
| HSES VS EO | 444.0 | 21.0 | **0.000013** |
| HSES VS AO | 465.0 | 0.0 | **0.000002** |
| HSES VS GSK | 230.0 | 205.0 | 0.778632 |
| HSES VS HGSA | 449.0 | 16.0 | **0.000008** |
| HSES VS IGOA | 465.0 | 0.0 | **0.000002** |
| HSES VS IMFO | 436.0 | 29.0 | **0.000027** |
| HSES VS MFLA | 346.0 | 89.0 | **0.005281** |
| HSES VS MPA | 375.0 | 90.0 | **0.003269** |
| HSES VS MSCA | 458.0 | 7.0 | **0.000003** |
| HSES VS SDCS | 465.0 | 0.0 | **0.000002** |
| HSES VS EBCM | 53.0 | 412.0 | **1** |
| ED-EB VS EO | 460.0 | 5.0 | **0.000003** |
| ED-EB VS AO | 465.0 | 0.0 | **0.000002** |
| ED-EB VS GSK | 388.0 | 47.0 | **0.000218** |
| ED-EB VS HGSA | 449.0 | 16.0 | **0.000008** |
| ED-EB VS IGOA | 465.0 | 0.0 | **0.000002** |
| ED-EB VS IMFO | 454.0 | 11.0 | **0.000005** |
| ED-EB VS MFLA | 386.0 | 49.0 | **0.000258** |
| ED-EB VS MPA | 426.0 | 39.0 | **0.000066** |
| ED-EB VS MSCA | 465.0 | 0.0 | **0.000002** |
| ED-EB VS SDCS | 458.0 | 7.0 | **0.000003** |
| ED-EB VS EBCM | 153.0 | 282.0 | 1 |
| LS-SPA VS EO | 460.0 | 5.0 | **0.000003** |
| LS-SPA VS AO | 465.0 | 0.0 | **0.000002** |
| LS-SPA VS GSK | 347.0 | 88.0 | **0.004939** |
| LS-SPA VS HGSA | 448.0 | 17.0 | **0.000009** |
| LS-SPA VS IGOA | 465.0 | 0.0 | **0.000002** |
| LS-SPA VS IMFO | 454.0 | 11.0 | **0.000005** |
| LS-SPA VS MFLA | 385.0 | 50.0 | **0.00028** |
| LS-SPA VS MPA | 410.0 | 55.0 | **0.000251** |
| LS-SPA VS MSCA | 461.0 | 4.0 | **0.000002** |
| LS-SPA VS SDCS | 458.0 | 7.0 | **0.000003** |
| LS-SPA VS EBCM | 184.0 | 251.0 | 1 |
| NLSHADE VS EO | 435.0 | 0.0 | **0.000002** |
| NLSHADE VS AO | 463.0 | 2.0 | **0.000002** |
| NLSHADE VS GSK | 417.0 | 18.0 | **0.000015** |
| NLSHADE VS HGSA | 460.0 | 5.0 | **0.000003** |
| NLSHADE VS IGOA | 462.0 | 3.0 | **0.000002** |
| NLSHADE VS IMFO | 459.0 | 6.0 | **0.000003** |
| NLSHADE VS MFLA | 450.0 | 15.0 | **0.000007** |
| NLSHADE VS MPA | 428.0 | 7.0 | **0.000005** |
| NLSHADE VS MSCA | 463.0 | 2.0 | **0.000002** |
| NLSHADE VS SDCS | 460.0 | 5.0 | **0.000003** |
| NLSHADE VS EBCM | 121.0 | 314.0 | 1 |

#### 3.3.1.3 Results of the CD plot

In the cases of the functions with 30 variables, EBCM exhibits similar performance to the 4 state-of-the-art algorithms in Fig.4, and the performance of GSK and MFLA are significantly similar to HSES, LS-SPA, and ED-EB. In addition, there was no significant difference between MPA and HSES. In contrast, the performance of the other 7 recent algorithms (i.e., AO, IMFO, EO, SDCS, HGSA,

MSCA, and IGOA) is significantly different from that of the 4 state-of-the-art algorithms. The conclusions drawn in this case are similar to the observation results of the Bayesian rank-sum test, the Friedman test, and the Wilcoxon signed-rank test.

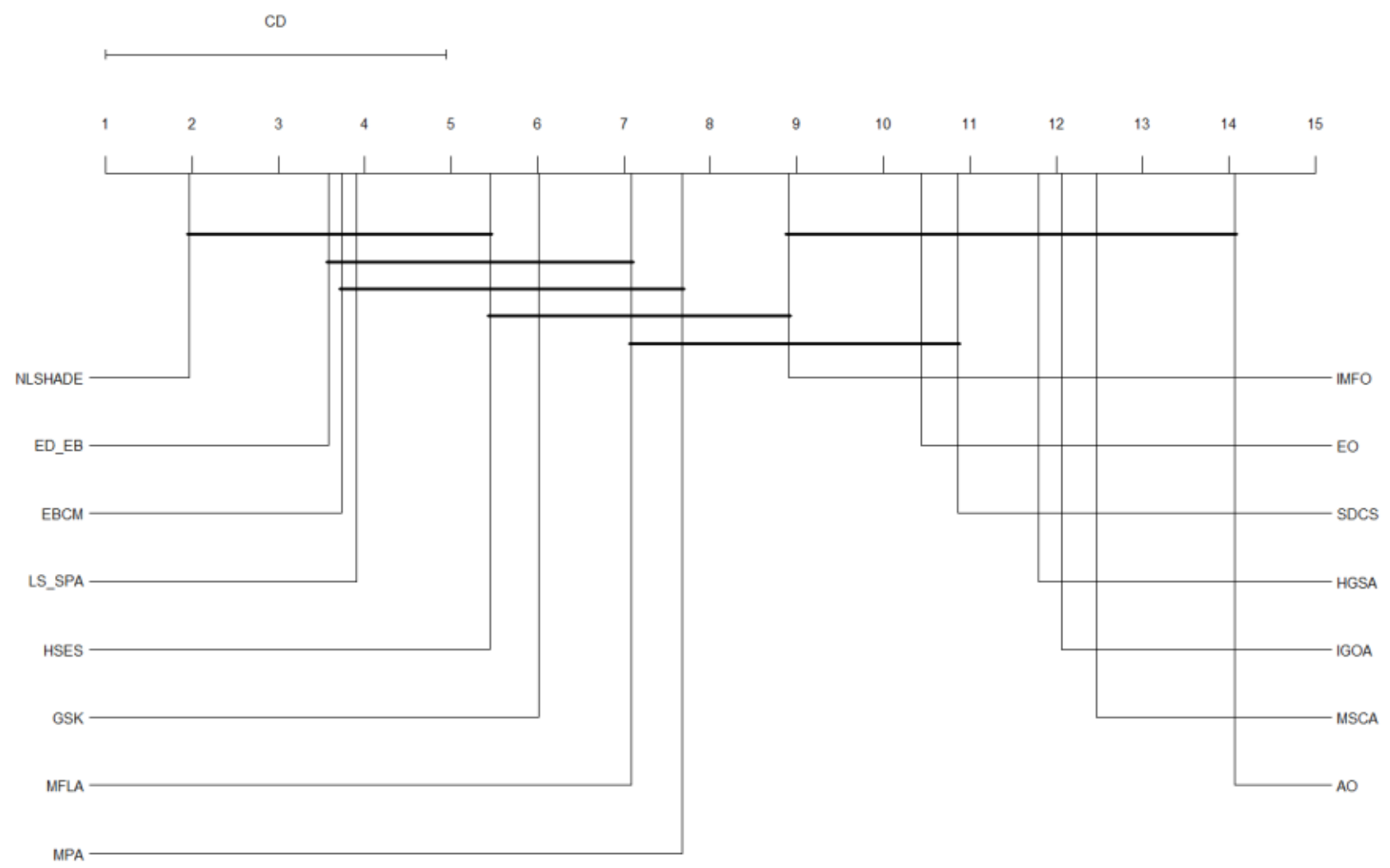

Fig.4 The CD plot of algorithms on the CEC 2017 functions with 30 variables

#### 3.3.1.4 Convergence analysis

The convergence plots of the 15 algorithms on functions F1, F3, F4, F10, F11, F19, F21 and F24 with 30 variables are shown in Fig.5. According to Fig.5, EBCM and 4 state-of-the-art algorithms have a fast convergence speed and can obtain better solutions on these selected functions compared with the other 10 new algorithms. Compared with the 4 state-of-the-art algorithms, AO, MFLA, HGSA, IGOA, and MSCA have a slower convergence speed and the worst global search ability (i.e., less efficient) on these selected functions. The other algorithms, such as EO, GAK, IMFO, MPA and SDCS, have a similar convergence speed to the 4 state-of-the-art algorithms, but they are inferior to the 4 state-of-the-art algorithms on most of these select functions. Particularly, there is a clear gap between the 4 state-of-the-art algorithms and the 11 new algorithms on functions F3, F7, F11, F21 and F24. These results suggest that the 4 state-of-the-art algorithms have a faster convergence speed and a stronger global search ability on most of the selected functions.

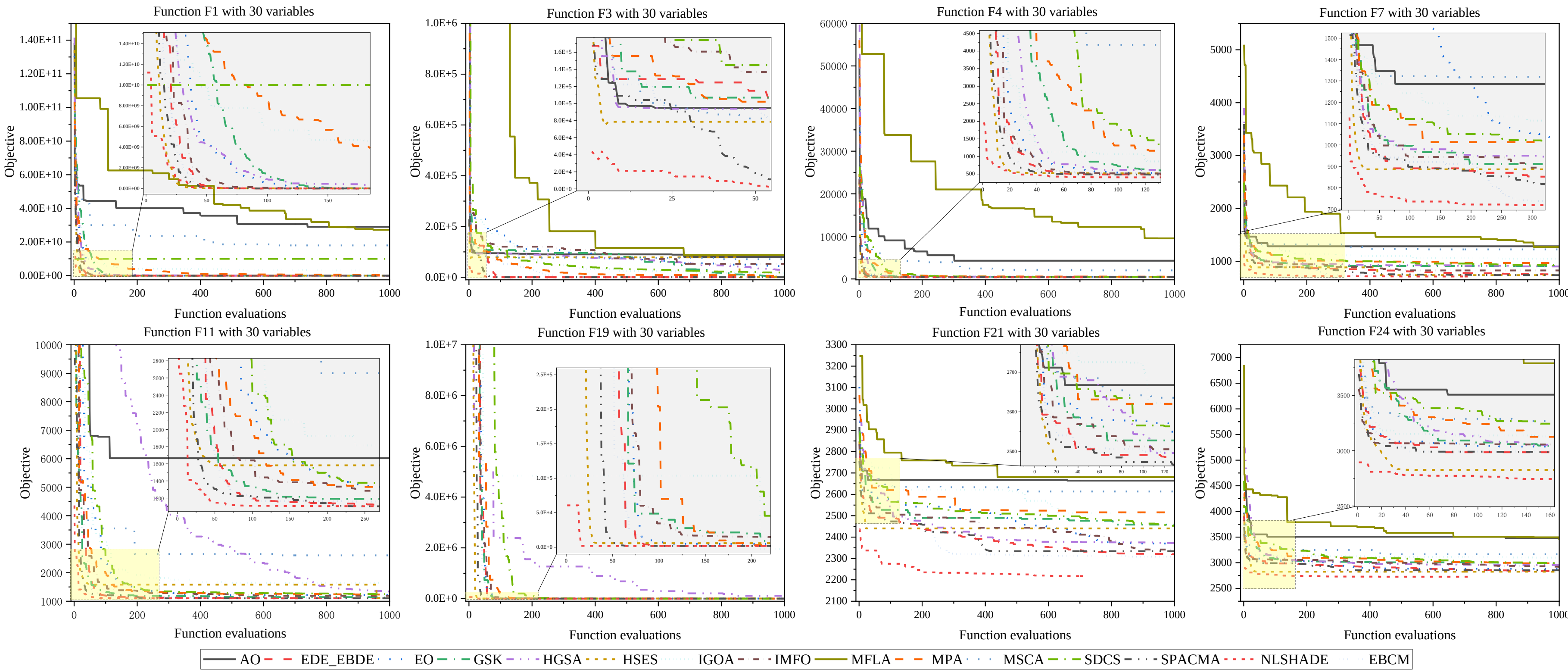

Fig.5 Convergence plots on functions with 30 variables

#### 3.3.1.5 The trade-off of exploration and exploitation analysis

We consider the method proposed in Ref. [93] to evaluate the trade-off between exploration and exploitation of the 15 algorithms. In particular, the percentage of exploration (i.e., XPL%) and the percentage of exploitation (i.e., XPT%) are used to evaluate the trade-off response. XPL% represents the level of exploration as a relationship between the diversity in each iteration and the maximum reached diversity. XPT% corresponds to the level of exploitation. Both elements, XPL% and XPT% are mutually conflicting and complementary. For more information about how to evaluate the trade-off of algorithms, please refer to [93].

Fig.S8 shows the experimental results of the exploration and exploitation trade-off of the 15 algorithms on functions F1, F3, F4, F7, F11, F19, F22 and F24 with 30 variables. Due to space limitations, we only analyze the results on functions F1, F7, F11 and F22. For more information about the trade-off analysis of the 15 algorithms on other functions, please read the supplementary material. The analysis results of Fig. S8 and Table S2 are summarized as follows.

- Unimodal function F1: In terms of the 4 state-of-the-art algorithms, LS-SPA and NLSHADE are the two most prominent algorithms on function F1, with an exploitation of 98.55% and 98.30%, and an exploration of 1.45% and 1.70%, respectively. ED-EB and HSES perform slightly worse than LS-SPA and NLSHADE and have trade-off behaviors of exploitation and exploration similar to these of two top algorithms. They exploited the search space 99.19% and 97.82% of the time, respectively. In terms of the 10 new algorithms, EBCM performs slightly worse than LS-SPA and NLSHADE but has similar trade-off exploitation and exploration behaviors as LS-SPA and NLSHADE. Meanwhile, the search space is exploited 96.95% of the time. MFLA, GSK, and IMFO also perform better among the 11 newly proposed algorithms, and they spent 85.58%, 99.13%, and 99.15% of the time exploiting the search space. Although MFLA employs a different exploration and exploitation rate compared with LS-SPA and NLSHADE, it benefits from multiple exploration peaks appearing during the optimization process to jump into different search zones and find better solutions. The other new algorithms including EO, AO, MPA, HGSA, IGOA, MSCA and SDCS, are less efficient in terms of solution quality and exploit the search space 96.78%, 93.31%, 98.84%, 99.56%, 79.76%, 66.20%, and 0.00% of the time, respectively. In particular. SDCS uses excessive exploration (i.e., 100% of the time) in its search process. EO, AO, MPA, and HGSA maintain a behavior very close to the one used by the top two algorithms but the reason for finding different solutions is because of the search mechanism used for exploration and exploitation. Moreover, MPA, IGOA and MSCA produce a very rough trade-off response. In all cases, the incremental-decremental graph shows that the exploration effect is very short, while the exploitation action is prolonged during most of the search time. The best trade-off can be found to be more than 98% exploitation and less than 2% exploration on function F1.
- Multimodal function F7: The 4 state-of-the-art algorithms and EBCM are the top five best-performing algorithms for solving function F7, where NLSHADE achieves the best results with 90.49% exploitation and 9.52% exploration. Moreover, HSES, ED-EB, and LS-SPA spent 98.20%, 93.69% and 96.26% of the time exploiting respectively. In terms of the 11 new algorithms, EBCM shows competitive performance with 89.52% exploitation and 110.49% exploration. EO, MPA, IMFO, and HGSA perform slightly worse than the 4 state-of-the-art algorithms but their trade-off levels between exploration and exploitation are close to the 4 state-of-the-art algorithms. The performance of AO, IGOA, MFLA, GSK, SDCS, and MSCA widens the gap with the 4 state-of-the-art algorithms and has different trade-off levels. They spent 97.01%, 78.41%, 83.98%, 92.54%, 98.40%, and 76.86% of the time exploring the search space, respectively. Particularly, IGOA, MFLA, and MSCA focus less on exploita-tion compared with the 4 state-of-the-art algorithms. On the contrary, SDCS has a slightly higher exploitation rate. Moreover, AO, GSK, and SDCS have similar trade-off levels between exploration and exploitation compared with the 4 state-of-the-art algorithms but have different solution qualities. Once again, it is a good example of how the difference in the quality of the specific search mechanism of each algorithm greatly affects the performance. According to the incremental-decremental graph, all 15 algorithms focused on exploitation, with a trade-off of more than 90% of the time exploring and less than 10% exploiting.

- Hybrid function F11: The top six best-performing algorithms for solving function F11 are the 4 state-of-the-art algorithms, EBCM and GSK. NLSHADE achieves the best results that exploited the search space 90.58% of the time and explored 9.42% of the time. HSES, ED-EB and LS-SPA exploited the search space 96.89%, 96.95%, and 95.14% of the time, respectively. GSK is slightly better than HSES and ED-EB with an exploitation of 95.03% and 4.97% exploration, respectively. Regarding the other new algorithms, EBCM obtains similar results to the 4 state-of-the-art algorithms and exploited the search space 88.99% of the time. MPA, MFLA, IMFO, and SDCS are inferior to the 4 state-of-the-art algorithms on function F11. They explored the search space 97.00%, 54.36%, 96.86%, and 97.04% of the time, respectively. The worst five algorithms are AO, EO, IGOA, HGSA, and MSCA, which achieve the exploitation of 53.31%, 97.88%, 79.10%, 99.59%, and 60.31%, respectively. These results show that MFLA, AO, and MSCA focused less on exploitation compared with the 4 state-of-the-art algorithms. In other words, their exploitation and exploration rates are not much different. In contrast, EO, MPA, and HGSA seem to focus slightly more on exploitation compared with the 4 state-of-the-art algorithms. Moreover, IGOA, HGSA and MSCA once again produce a rough trade-off response, and always seem inefficient. The incremental-decremental graph shows that the best-performing algorithms that prefer exploitation to exploration, and are closer to 90% exploitation and 10% exploration, are used in their search process.
- Composition function F22: The results in Table S2 suggest that the most prominent algorithms for solving function F22 are 4 state-of-the-art algorithms and EBCM. In addition, GSK, MFLA, and SDCS are the distant seconds. In terms of the 4 state-of-the-art algorithms, HSES, ED-EB, and LS-SPA exploited the search space 98.08%, 98.87%, and 98.55 of the times, respectively. On the contrary, they spent 1.92%, 1.13%, and 1.45% of their time exploring, respectively. NLSHADE focused less on exploitation than the other four state-of-the-art algorithms, with an exploitation of 96.82% and an exploration of 3.182%. In terms of the 11 new algorithms, EBCM attains similar results to the 4 state-of-the-art algorithms, which exploited the search space 99.27% of the time and explored 0.735% of the time. GSK and MFLA have a better performance that achieves an exploitation of 97.57% and 81.69%, respectively. AO, MPA, SDCS, MSCA, and HGSA are slightly inferior to the seven top algorithms and have different trade-off levels. They spent 87.53%, 97.82%, 99.00%, 59.21% and 99.54% on exploitation, respectively. Moreover, EO, IGOA, and IMFO are the three worst algorithms for solving function F22. Although EO, GSK, MPA, HGSA, and SDCS obtain trade-off levels that are very similar to those of the best seven algorithms, they present bad performance in terms of the solution quality. This once again shows the importance of the search mechanisms to obtain a better performance. It is important to note that the 4 state-of-the-art algorithms, GSK and SDCS, produce the smoothest trade-off response, but EO, MPA, HGSA, IGOA, MPLA, and MSCA produce a rough trade-off response. According to the incremental-decremental graph, all the algorithms focused more time on exploitation, and the best trade-off for function F22 is closer to 99% exploitation and 1% exploration.

In summary, EBCM has a similar performance and trade-off behavior of exploitation and explora-tion compared with the 4 state-of-the-art algorithms. Furthermore, GSK, MPA, MFLA, and IMFO are slightly inferior to the 4 state-of-the-art algorithms but demonstrate better performance than the other 6 new algorithms (i.e., except for EBCM). Although each algorithm has different exploitation and exploration behaviors on each function, all the algorithms focus more time on exploitation, especially the better-performing algorithms. Due to space limitations, we only show the balancing behavior of GSK and EBCM on functions F1, F7, F11 and F22, as shown in Fig.6.

**3.3.1.6 Diversity analysis**

To complement the analysis, an experiment of diversity on functions F3, F7, F11 and F24 with 30 variables is conducted and the results are presented in Fig. 7. In the experiments, we consider Equations (1) and (2) defined in Ref. [93] for a diversity assessment and these two equations are shown below.

$$Div_j = \frac{1}{n}\sum_{i=1}^{n} |median(x^j) - x_i^j| \tag{42}$$

$$Div = \frac{1}{m}\sum_{j=1}^{m} Div_j \tag{43}$$

where $median(x^j)$ represents the median of dimension $j$ in the whole population. $x_{ij}$ is the dimension $j$ of search agent $i$. $n$ corresponds to the number of search agents in the population while $m$ symbolizes the number of design variables of the optimization problem.

According to Fig.7, it is clear that all 13 algorithms (i.e., except for AO and MSCA) begin with a large diversity as a consequence of their random initialization. As the number of iterations increase, the population diversity diminishes. AO and MSCA also begin with a large diversity but they have a certain population diversity at the final stage of iteration. Especially the diversity of AO on functions F11 and F24 first decreases and then increases gradually with the iterations. Most of the 11 new algorithms show a rough trade-off response, especially MPA, MSCA, MFLA, IGOA and HGSA, which exhibit high oscillation behavior. Compared with the 11 new algorithms, the 4 state-of-the-art algorithms show the smoothest diversity responses.

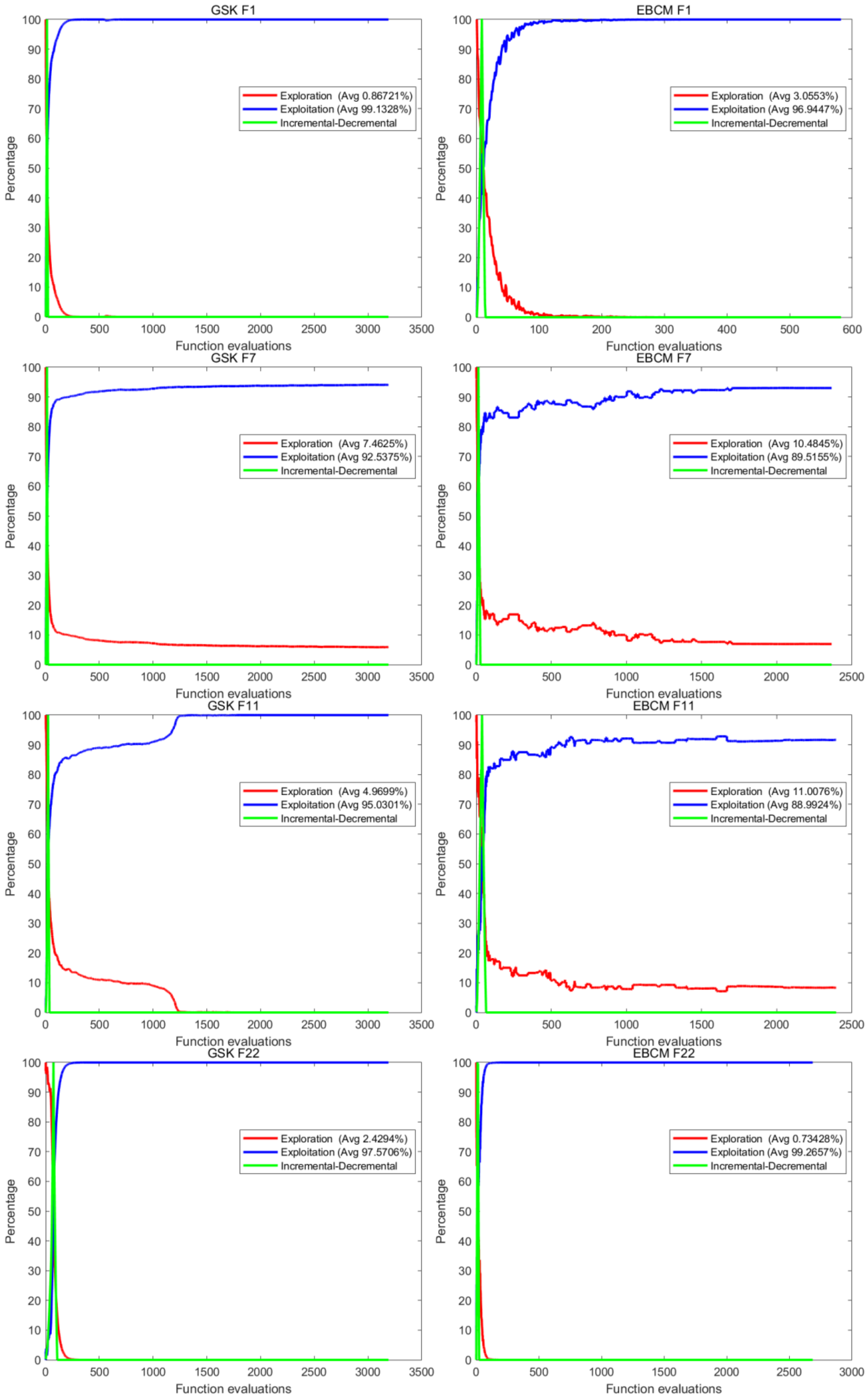

Fig.6 The balancing behavior of GSK and EBCM on functions F1, F7, F11, and F22

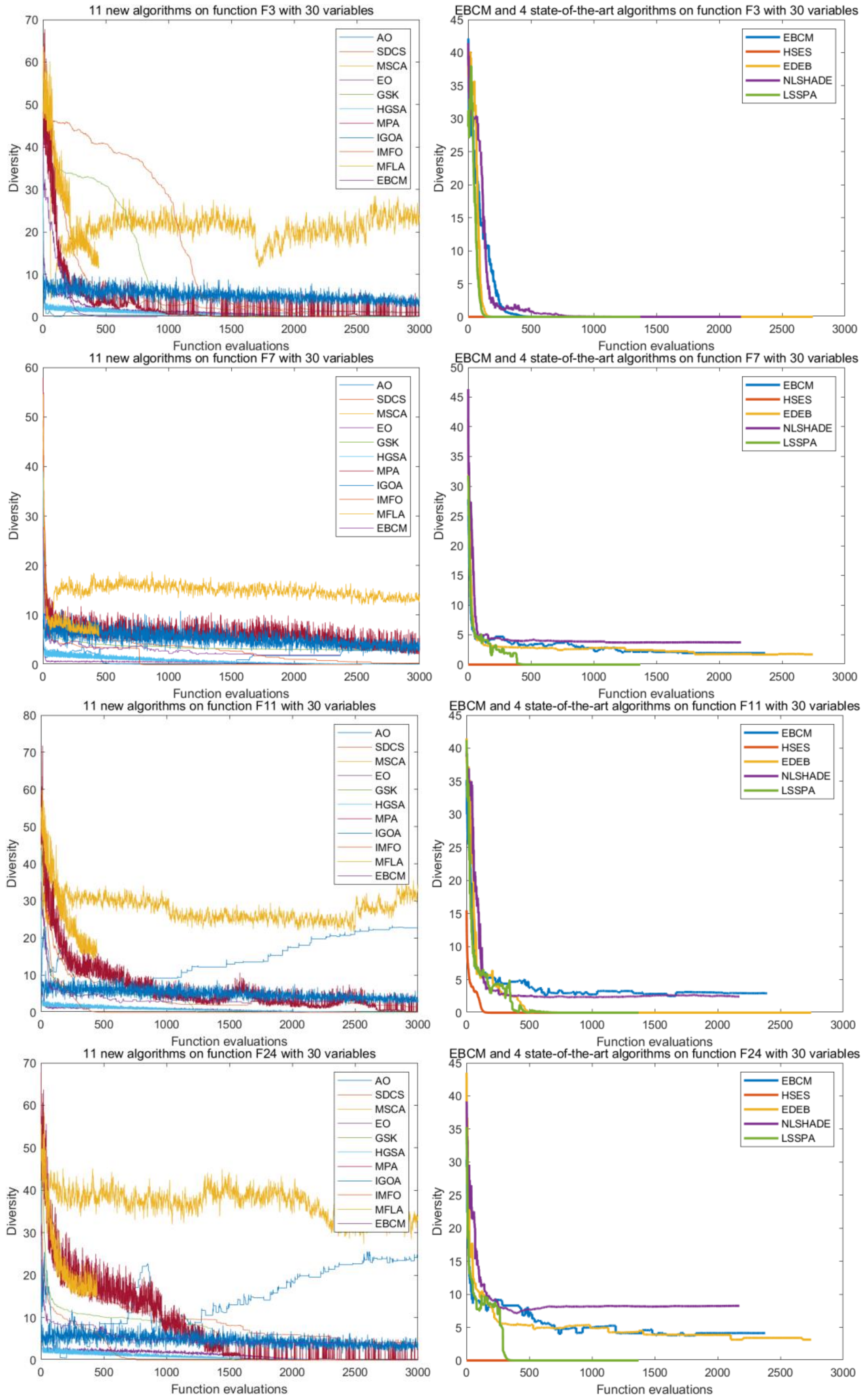

Fig.7 Diversity analysis on functions F3, F7, F11, and F24 with 30 variables

### 3.3.2 Results of CEC 2017 functions considering nonshifted and shifted

#### 3.3.2.1 Evaluate the search bias toward the origin

In the literature, some algorithms perform well when solving problems whose optimal solutions are located at the origin/center of the search space, but they are less efficient when dealing with the same functions whose optimal solutions deviate from the origin. Liang et al. [124] first evaluated the performance of the multiagent genetic algorithm by considering the searches biased to the origin. In addition, some newly proposed algorithms, such as TLBO [22,23] and GWO [24], have also been verified to have a search bias to the origin. In this section, extensive experiments are carried out on the shifted and non-shifted CEC 2017 functions and consider 10 and 30 variables to evaluate whether the 15 algorithms (11 recent algorithms and 4 state-of-the-art algorithms) have a search bias to the origin. Detailed information on the experimental results is shown in Tables S23-S24. Nonparametric methods, such as the Friedman test and the Wilcoxon signed-ranks test are used to further analyze the experimental data.

The results of the Friedman test in Table 4 show that SDCS, MSCA, MFLA, and MPA have a better performance compared with the 4 state-of-the-art algorithms for solving the nonshifted and shifted functions with 10 and 30 variables. Particularly, the SDCS achieves the lowest rank with scores of 3.4 and 3.6, respectively. Compared with the results obtained on the shifted functions, the performance of AO, SDCS, MSCA, and MFLA are significantly improved for solving the nonshifted functions with 10 and 30 variables. The performance of SDCS, MSCA and MFLA are significantly affected by the shift operator on the functions. In other words, these algorithms have search that are biased to the origin. The performance of the other recent algorithms (i.e., EO, IGOA, and HGSA) is slightly improved for solving the nonshifted functions compared with the algorithms dealing with the shifted functions. In addition, the performance of the 4 state-of-the-art algorithms has no significant difference in the solution between the nonshifted functions and the shifted functions.

The results of the Wilcoxon Signed Ranks test are shown in Tables 5-6. It can be found that the values of $R^+$ are generally greater than the values of $R^-$, which means that the 15 algorithms can obtain better performances in solving the nonshifted functions. In other words, all 15 algorithms have searches biased to the origin and are affected to varying degrees. For example, the 4 state-of-the-art algorithms are less affected by the shift operator on the functions compared with the 11 recent algorithms. In contrast, AO, SDCS, MFLA, MSCA, HGSA, and IGOA among the 11 recent algorithms are greatly affected. In addition, we draw the CD plots for the experimental results as shown in Figs.8-9, which have similar conclusions to the observations from the Friedman test and the Wilcoxon signed ranks test.

In conclusion, all 15 comparative algorithms suffer from search biases to the origin to varying degrees. In particular, the 4 state-of-the-art algorithms are less affected by the shift operator on the functions compared with the 11 recent algorithms.

Table 4 The results of the Friedman test on the nonshifted and shifted CEC2017 functions

| Algorithms | Average ranking on functions with 10 variables | | Average ranking on functions with 30 variables | |
|---|---|---|---|---|
| | nonshifted | shifted | nonshifted | shifted |
| EO | 9.3333 (9) | 11.0333 (12) | 8.6 (8) | 8.1167 (10) |
| MPA | 5.9833 (5) | 3.85 (1) | 6.3667 (6) | 7.55 (9) |
| GSK | 10.1 (11) | 7.75 (9) | 11.2333 (13) | 7.3667 (4) |
| MSCA | 4.2833 (2) | 10.4333 (11) | 4.35 (3) | 9.55 (14) |
| IMFO | 11.6167 (14) | 8.7667 (10) | 13.7667 (15) | 7.5 (8) |
| MFLA | 4.5333 (3) | 5.5667 (5) | 4.3833 (4) | 7.3667 (4) |
| HGSA | 11.75 (15) | 12 (13) | 12.3833 (14) | 8.5333 (12) |
| IGOA | 11.6 (13) | 12.4667 (14) | 9.4833 (11) | 9.0833 (13) |
| SDCS | 3.9333 (1) | 7.65 (8) | 4 (2) | 8.25 (11) |
| AO | 5.0333 (4) | 13.8667 (15) | 3.8167 (1) | 9.85 (15) |
| NLSHADE | 9.3833 (10) | 5.7 (6) | 6.3 (5) | 7.3667 (4) |
| EBCM | 7.7667 (7) | 4.1833 (2) | 8.733 (9) | 7.3667 (4) |
| HSES | 10.1667 (12) | 7.4333 (7) | 10.1667 (12) | 7.3667 (4) |
| LS-SPA | 6.4333 (6) | 4.75 (4) | 7.45 (7) | 7.3667 (4) |
| ED-EB | 8.0833 (8) | 4.55 (3) | 8.9167 (10) | 7.3667 (4) |

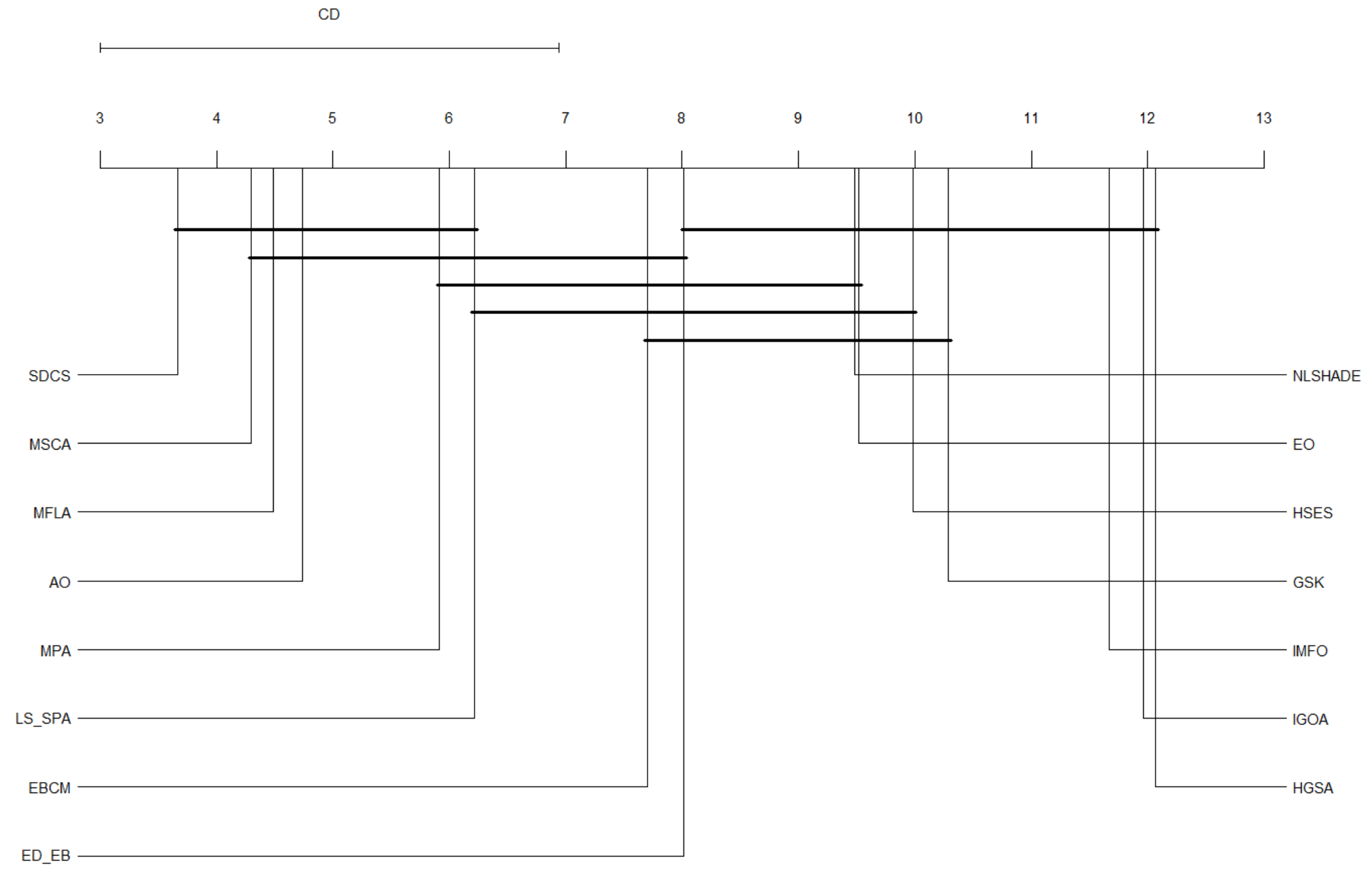

Fig.8 The CD plot of algorithms on the nonshifted functions with 10 variables

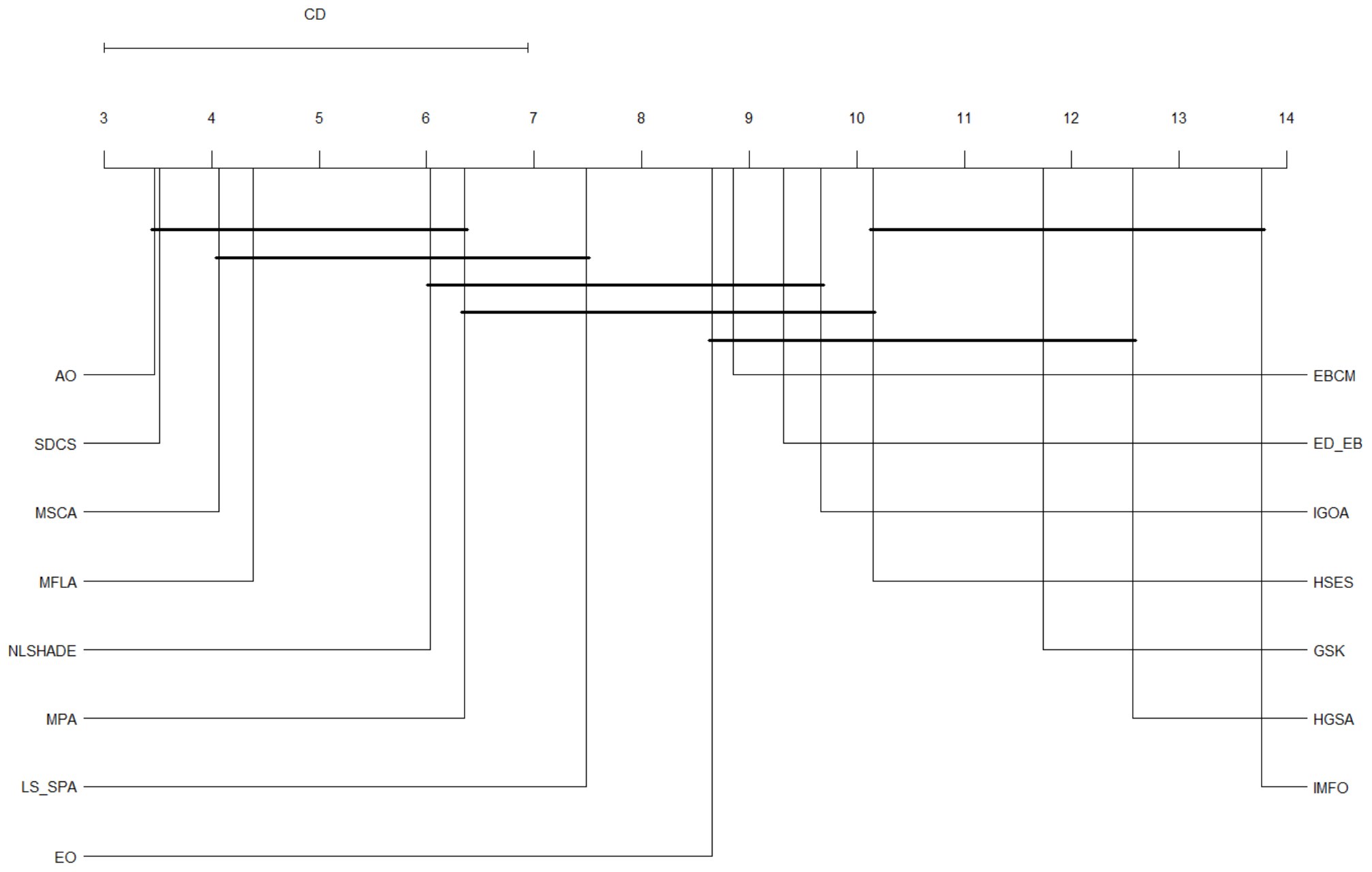

Fig.9 The CD plot of algorithms on the nonshifted functions with 30 variables

**3.3.2.2 The trade-off response of AO, MFLA, MSCA, and SDCS on the nonshifted functions**

Because AO, MFLA, MSCA, and SDCS maintain good performance in problems where the optimal point is in the origin, we conduct experiments to investigate the trade-off response of these four algorithms on functions F7 and F24 with 10 and 30 variables, and the results are presented in Figures S13-S14. As can be seen from Fig.S13, the evolution of the trade-off in the nonshifted functions for AO demonstrates consistency with the trade-off made by them on the shifted functions. MFLA focuses slightly less on exploration when considering functions that are shifted. It exploited the search space 83.53% of the time on the nonshifted function F7 while 72.68% of the time exploiting on the shifted function F7. Moreover, MFLA spent 46.87% and 42.70% of the time exploiting the search space of the nonshifted and shifted function F24, respectively. In contrast, MSCA and SDCS focused slightly more exploration on the nonshifted functions than the shifted functions. According to Fig.S14, MFLA, MSCA, and SDCS present consistent trade-off responses between nonshifted and shifted functions. The case of AO on function F24 is interesting. AO produces a rougher trade-off response on the shifted F24 than on the nonshifted F24. In particular, AO employed a trade-off of 96.91% exploitation and 3.09% exploration on the problem where the optimal solution is in the origin. On the shifted function F24, AO spent 76.94% on exploiting and 23.07% on exploring. Especially the SDCS maintains the same trade-off response between nonshifted and shifted. Nevertheless, its performance in terms of quality is significantly better than the nonshifted functions. This seems to indicate that the SDCS maintains a fixed trade-off without considering the function type. In a word, it is important to point out that these four algorithms present a rough trade-off response on non-shifted and shifted functions F7 and F24. These results suggest that the search mechanisms used by these algorithms seem to have a significant impact on their performance, which we will investigate in our future work.

Table 5 The results of Wilcoxon signed-ranks test on the shifted and non-shifted functions with 10 variables

| No. | EO-nonshifted VS EO-shifted | AO-nonshifted VS AO-shifted | GSK-nonshifted VS GSK-shifted | HGSA-nonshifted VS HGSA-shifted | IGOA-nonshifted VS IGOA-shifted | IMFO-nonshifted VS IMFO-shifted | MFLA-nonshifted VS MFLA-shifted | MPA-nonshifted VS MPA-shifted | MSCA-nonshifted VS MSCA-shifted | SDCS-nonshifted VS SDCS-shifted | HSES-nonshifted VS HSES-shifted | EBCM-nonshifted VS EBCM-shifted | ED-EB-nonshifted VS ED-EB-shifted | LS-SPA-nonshifted VS LS-SPA-shifted | NLSHADE-nonshifted VS NLSHADE-shifted |
|---|---|---|---|---|---|---|---|---|---|---|---|---|---|---|---|
| $R^+$ | 337.5 | 465.0 | 278.5 | 426.0 | 388.0 | 270.0 | 400.5 | 346.5 | 461.0 | 423.5 | 274.0 | 308.0 | 281.5 | 375.5 | 309.0 |
| $R^-$ | 97.5 | 0.0 | 186.5 | 39.0 | 77.0 | 165.0 | 64.5 | 118.5 | 4.0 | 41.5 | 161.0 | 127.0 | 153.5 | 89.5 | 126.0 |
| $p$-value | 0.0083 | 0.000002 | >=0.2 | 1.3966E-5 | 8.718E-4 | >=0.2 | 2.697E-4 | 0.01799 | 1.3038E-8 | 1.9522E-5 | >=0.2 | 0.05064 | 0.1723 | 0.002464 | >=0.2 |

Table 6 The results of Wilcoxon signed-ranks test on the shifted and non-shifted functions with 30 variables

| No. | EO-nonshifted VS EO-shifted | AO-nonshifted VS AO-shifted | GSK-nonshifted VS GSK-shifted | HGSA-nonshifted VS HGSA-shifted | IGOA-nonshifted VS IGOA-shifted | IMFO-nonshifted VS IMFO-shifted | MFLA-nonshifted VS MFLA-shifted | MPA-nonshifted VS MPA-shifted | MSCA-nonshifted VS MSCA-shifted | SDCS-nonshifted VS SDCS-shifted | HSES-nonshifted VS HSES-shifted | EBCM-nonshifted VS EBCM-shifted | ED-EB-nonshifted VS ED-EB -shifted | LS-SPA-nonshifted VS LS-SPA -shifted | NLSHADE-nonshifted VS NLSHADE -shifted |
|---|---|---|---|---|---|---|---|---|---|---|---|---|---|---|---|
| $R^+$ | 415.0 | 465.0 | 332.0 | 419.0 | 462.0 | 223.0 | 392.0 | 394.0 | 465.0 | 452.0 | 352.0 | 338.0 | 337.5 | 343.0 | 289.0 |
| $R^-$ | 50.0 | 0.0 | 133.0 | 46.0 | 3.0 | 242.0 | 43.0 | 71.0 | 0.0 | 13.0 | 113.0 | 97.0 | 97.5 | 122.0 | 146.0 |
| $p$-value | 5.592E-5 | 0.000002 | 0.04048 | 3.454E-5 | 9.314E-9 | >=0.2 | 4.7E-5 | 5.054E-4 | 1.8626E-9 | 1.6392E-7 | 0.012834 | 0.008008 | 0.008301 | 0.0221 | >=0.2 |

## 4 Issues and suggestions for future research

Despite the fruitful results of metaheuristic research in the past few decades, there are still some suggestions and interesting open problems that need to be investigated in future research.

- **Fair and comprehensive comparisons:** For a fair comparison, it is necessary to configure the parameters of all the comparative algorithms using the same satisfactory parameter tuning approach, as the performance of metaheuristics is severely affected by the parameter settings. In addition, when evaluating the performance of a newly proposed algorithm, it is mandatory to compare with the state-of-the-art variants of established metaheuristics on comprehensive and representative benchmark suits with rotated and shifted problems. Almost all the metaheuristic algorithms are stochastic, which means that they may obtain results of different quality in different runs. Therefore, rigorous statistical tests are useful in comparing different metaheuristic algorithms [125,126]. In some cases, the details of an algorithm are not fully explained due to space limitations, which may result in inaccurate replication and inconsistent computation results. Thus, it is highly recommended that the authors make the source codes publicly available.
- **Improve and propose metaheuristics from search behavior and optimization mechanisms perspectives:** Recent metaheuristics are proposed according to phenomena from biology, nature, physics, and so on. However, the effective performance of metaheuristics essentially depends on the search behaviors and optimization mechanisms. For example, the neighborhood structures (e.g., one-point exchange and multiple-point exchange) in single-solution based metaheuristics, and the operators (e.g., crossover, mutation, and recombination) in population-solution based metaheuristics play crucial roles in the high performance of the optimizers. Besides, how to guide the search directions to a promising region in the solution space is another promising optimization mechanism. In particular, balancing exploration and exploitation to improve the performance of metaheuristics is significant. Therefore, we appeal to researchers to improve and propose metaheuristics not only from the inspiration source but also from the perspectives of the essential search behaviors and optimization mechanisms.
- **Automatic design and configuration of metaheuristics:** The design and configuration of metaheuristics can be considered an optimization problem. Traditional methods depend on prior knowledge and trial-and-error methods to obtain a configuration. Automatic design and configuration methods are attracting attention in the fields of metaheuristics [127]. It not only saves a substantial amount of human effort during the empirical analysis and design of metaheuristics but also leads to high-performance optimizers [128]. Therefore, it is worth using automatic methods to design and configure metaheuristics. For example, there are many operators, neighborhood structures, parameters, and mechanisms of information sharing and learning in the component pool. These components may be adaptively automatically selected from the component pool based on the features of the problems, and can be effectively combined to design efficient algorithms for solving the specific problems. With regard to metaheuristic design, LaTorre et al. [117] suggested that simplicity should be considered one of the preferential aspects in the design of new optimization techniques. Particularly, some new algorithms are improved on previous algorithms by updating or adding new strategies to their search procedure. Each improvement/component that affects the performance of the new algorithm needs to be further analyzed [129].
- **Combining machine learning techniques with metaheuristics:** Machine learning (ML) has achieved fruitful results in recent decades. ML's powerful learning, prediction, and decision-making capabilities have opened a new horizon for metaheuristic research. It is promising to combine

ML and metaheuristics in the following aspects: 1) A combination of meta-heuristics and deep learning, reinforcement learning, ensemble learning, etc., and reasonable recommendation of optimization algorithms for specific problems [130,131]. 2) Using ML techniques to help to model optimization problems, analyze the solution space, and perform problem decomposition [132,133]. 3) ML can use historical data to dynamically adjust parameter values during the optimization process of metaheuristics. Besides, when metaheuristics have multiple operators and search mechanisms, ML is a prevalent and effective method for learning the characteristics of these operators and mechanisms, and for generating the appropriate algorithmic configuration [134,135].

- **Integrate problem domain knowledge into metaheuristics:** Integrating algorithms with problem domain knowledge can improve the performance of the algorithms. For instance, designing the operators and search mechanisms of metaheuristics based on the problem characteristics leads to having the search directions of the algorithms based on the landscapes of the problem. In addition, the optimality conditions of the problems can also be used to reduce the variables and the difficulty of the problems considered [136].
- **Application to complex real-world optimization problems:** Most real-world optimization problems are large-scale, with complex constraints, high-dimensional objectives, continuous variables and discrete variables. However, metaheuristics also face quite a few challenges when solving these complex real-world optimization problems. It is efficient to combine metaheuristics with surrogate models [1] such as parallel acceleration and simulation optimization to solve complex real-world optimization problems.

# 5 Conclusions

In this paper, we provide a comprehensive review of metaheuristics. More than 500 newly proposed and improved metaheuristics are collected, and a taxonomy of metaheuristics is further proposed to describe the metaheuristics from two aspects, including the inspiration sources and the essential operators for generating solutions. We find that the recent metaheuristics proposed in the literature are neither rigorously tested on comprehensive and representative benchmark suites, nor compared with the state-of-the-art metaheuristics. Therefore, to evaluate and understand the performance of the state-of-the-art and recent metaheuristics, 11 representative metaheuristics with new names are selected as recent algorithms to be compared with the 4 state-of-the-art algorithms on the CEC 2017 benchmark suite.

For fair comparisons, we first use a unified framework named irace to automatically configure the parameters of all 15 comparative algorithms. Then, whether these algorithms have a search bias to the origin is investigated. For a detailed description, the convergence, diversity and trade-off between the exploration and exploitation of all 15 algorithms are also analyzed. All the experimental results were analyzed by nonparametric statistical methods, including the Friedman test, Wilcoxon signed-ranks test, and Bayesian signed-rank test. The results show that the performance of EBCM is similar to the 4 compared algorithms, and has the same properties and behaviors, such as convergence, diversity, exploration and exploitation trade-off, etc. But the other 10 recent algorithms are inferior to the 4 state-of-the-art algorithms for solving the CEC 2017 benchmark suite with 10, 30, and 50 variables. Besides, all 15 algorithms have searches biased to the origin but with different strengths. However, the 4 state-of-the-art algorithms are less affected by the shift operator of the functions compared with the 11 recent algorithms. Except for EBCM, it should be noted that the other 10 new algorithms are inferior to the 4 state-of-the-art algorithms in terms of convergence speed and global search ability on CEC 2017 func-tions. Moreover, the other 10 new algorithms show a rougher trade-off and diversity compared to the 4 state-of-the-art algorithms. Finally, several issues and suggestions based on the abovementioned review and experiments are proposed.

In the next part of this survey series, we extend our work from the following aspects:

(1) Metaheuristics are a broad field of research. We need to focus on comparative studies including examining more newly proposed algorithms and state-of-the-art algorithms on benchmarks and real-world problems of different sizes, complexes, and categories.

(2) Due to space limitations, we investigated the performance and properties of different metaheuristics in the current study. We need a thorough theoretical analysis to confirm why these metaheuristics perform better or worse.

(3) We investigated some metaheuristics on whether their search is biased toward the origin. However, which parameters or strategies influence this property requires further study.

Overall, we hope that our study provides useful insight to guide future designs of more practicable metaheuristics that are capable of handling complex, high-dimensional and large-scale real-world problems.

Table 7 List of metaheuristics

(This list will be posted at: https://github.com/P-N-Suganthan/MHA-500Plus)

| Ref | Year | Full name & abbreviation | Ref | Year | Full name & abbreviation |
|---|---|---|---|---|---|
| B1 | 1960 | Evolutionary Programming, EP | B257 | 2016 | Water Evaporation Optimization, WEO |
| B2 | 1964 | Evolution Strategies, ES | B258 | 2016 | Root Tree Optimization Algorithm, RTO |
| B3 | 1971 | Genetic Algorithm, GA | B259 | 2016 | FIFA World Cup Algorithm, FIFAWC |
| B4 | 1977 | Scatter Search Algorithm, SSA | B260 | 2016 | Sperm Whale Algorithm, SWA |
| B5 | 1981 | Genetic Programming, GP | B261 | 2016 | Virus Optimization Algorithm, VOA |
| B6 | 1983 | Simulated Annealing, SA | B262 | 2016 | Duelist Algorithm, DA |
| B7 | 1986 | Tabu Search Algorithm, TSA | B263 | 2016 | Raven Roosting Optimization Algorithm, RROA |
| B8 | 1989 | Stochastic Search Network, SSN | B264 | 2016 | Ring Seal Search, RSS |
| B9 | 1989 | Memetic Algorithm, MA | B265 | 2016 | Flying Elephant Algorithm, FEA |
| B10 | 1992 | Ant Colony Optimization, ACO | B266 | 2016 | Camel Algorithm, CA |
| B11 | 1993 | Shuffled Complex Evolution, SCE | B267 | 2016 | Crystal Energy Optimization Algorithm, CEO |
| B12 | 1993 | Great Deluge Algorithm, GDA | B268 | 2016 | Passing Vehicle Search, PVS |
| B13 | 1994 | Cultural Algorithms, CA | B269 | 2016 | Tug Of War Optimization, TWO |
| B14 | 1995 | Differential Evolution, DE | B270 | 2016 | Dynamic Virtual Bats Algorithm, DVBA |
| B15 | 1995 | Particle Swarm Optimization, PSO | B271 | 2016 | Lion Optimization Algorithm, LOA |
| B16 | 1995 | Old Bachelor Acceptance, OBA | B272 | 2016 | Natural Forest Regeneration Algorithm, NFR |
| B17 | 1996 | Bacterial Evolutionary Algorithm, BEA | B273 | 2016 | Simulated Kalman Filter, SKF |
| B18 | 1997 | Variable Neighbourhood Descent Algorithm, VND | B274 | 2016 | Shuffled Multi-Swarm Micro-Migrating Birds Optimization, $SM^2$-MBO |
| B19 | 1998 | Bee System, BS1 | B275 | 2016 | Yin-Yang-Pair Optimization, YYPO |
| B20 | 1998 | Photosynthetic Learning Algorithm, PLA | B276 | 2016 | Virulence Optimization Algorithm, VOA |
| B21 | 1998 | Chaos Optimization Algorithm, COA | B277 | 2017 | Artificial Butterfly Optimization, ABO |
| B22 | 1999 | Sheep Flocks Heredity Model, SFHD | B278 | 2017 | Cyclical Parthenogenesis Algorithm, CPA |
| B23 | 1999 | Extremal Optimization, EO | B279 | 2017 | Deterministic Oscillatory Search, DOS |
| B24 | 1999 | Gravitational Clustering Algorithm, GCA | B280 | 2017 | Fractal-based Algorithm, FA |
| B25 | 2000 | Clonal Selection Algorithm, CSA | B281 | 2017 | Neuronal Communication Algorithm, NCA |
| B26 | 2001 | Harmony Search Algorithm, HSA | B282 | 2017 | Lightning Attachment Procedure Optimization, LAPO |
| B27 | 2001 | Gene Expression Programming, GEP | B283 | 2017 | Bison Behavior Algorithm, BBA |
| B28 | 2001 | Marriage In Honey Bees Optimization, MBO | B284 | 2017 | Drone Squadron Optimization, DSO |
| B29 | 2002 | Bacterial Foraging Algorithm, BFA | B285 | 2017 | Human Behavior-Based Optimization, HBO |
| B30 | 2002 | Bacteria Chemotaxis Algorithm, BCA | B286 | 2017 | Vibrating Particles System, VPS |
| B31 | 2002 | Bee System, BS2 | B287 | 2017 | Spotted Hyena Optimizer, SHO |
| B32 | 2002 | Popmusic Algorithm, POPMUSIC | B288 | 2017 | Salp Swarm Algorithm, SSA |
| B33 | 2002 | Social Cognitive Optimization, SCO | B289 | 2017 | Grasshopper Optimisation Algorithm, GOA |
| B34 | 2003 | Artificial Fish Swarm Algorithm, AFSA | B290 | 2017 | Rain Fall Optimization Algorithm, RFO |
| B35 | 2003 | Covariance Matrix Adaptation–Evolution Strategy, CMA-ES | B291 | 2017 | Hydrological Cycle Algorithm, HCA |
| B36 | 2003 | Society and Civilization, SC | B292 | 2017 | Killer Whale Algorithm, KWA |
| B37 | 2003 | Artificial Immune System, AIS | B293 | 2017 | Camel Herd Algorithm, CHA |
| B38 | 2003 | Queen-bee Evolution, QBE | B294 | 2017 | Collective Decision Optimization Algorithm, CDOA |
| B39 | 2003 | Electromagnetism-Like Mechanism Optimization, EMO | B295 | 2017 | Laying Chicken Algorithm, LCA |
| B40 | 2004 | Beehive Algorithm, BHA | B296 | 2017 | Kidney-Inspired Algorithm, KIA |
| B41 | 2004 | Self-Organizing Migrating Algorithm, SOMA | B297 | 2017 | Golden Sine Algorithm, Gold-SA |
| B42 | 2005 | Artificial Bee Colony Algorithm, ABCA | B298 | 2017 | Sperm Motility Algorithm, SMA |
| B43 | 2005 | Bee Colony Optimization, BCO | B299 | 2017 | Rain Water Algorithm, RWA |
| B44 | 2005 | Bees Swarm Optimization Algorithm, BSOA | B300 | 2017 | Thermal Exchange Optimization, TEO |
| B45 | 2005 | Dendritic Cells Algorithm, DCA | B301 | 2017 | Porcellio Scaber Algorithm, PSA |
| B46 | 2005 | The Bees Algorithm, BA | B302 | 2017 | Selfish Herd Optimizer, SHO |
| B47 | 2005 | Wasp Swarm Optimization, WSO | B303 | 2017 | Polar Bear Optimization Algorithm, PBO |
| B48 | 2006 | Shuffled Frog-Leaping Algorithm, SFLA | B304 | 2017 | Social Engineering Optimization, SEO |
| B49 | 2006 | Big Bang–Big Crunch, BBC | B305 | 2017 | Sonar Inspired Optimization, SIO |

| Ref | Year | Full name & abbreviation | Ref | Year | Full name & abbreviation |
|---|---|---|---|---|---|
| B50 | 2006 | Cat Swarm Optimization, CSO | B306 | 2017 | Weighted Superposition Attraction, WSA |
| B51 | 2006 | Flocking base Algorithm, FA | B307 | 2017 | Satin Bowerbird Optimizer, SBO |
| B52 | 2006 | Honey-bees Mating Optimization Algorithm, HBMO | B308 | 2018 | Artificial Atom Algorithm, A3 |
| B53 | 2006 | Small-World Optimization Algorithm, SWOA | B309 | 2018 | Artificial Swarm Intelligence, ASI |
| B54 | 2006 | Saplings Growing Up Algorithm, SGUA | B310 | 2018 | Bees Life Algorithm, BLA |
| B55 | 2006 | Seeker Optimization Algorithm, SOA | B311 | 2018 | Beetle Swarm Optimization Algorithm, BSOA |
| B56 | 2006 | Weed Colonization Optimization, WCO | B312 | 2018 | Brunsvigia Optimization Algorithm, BVOA |
| B57 | 2007 | Imperialist Competitive Algorithm, ICA | B313 | 2018 | Car Tracking Optimization Algorithm, CTOA |
| B58 | 2007 | Monkey Search Algorithm, MSA | B314 | 2018 | Cheetah Based Algorithm, CBA |
| B59 | 2007 | River Formation Dynamics, RFD | B315 | 2018 | Cheetah Chase Algorithm, CCA |
| B60 | 2007 | Bacterial Swarming Algorithm, BSA | B316 | 2018 | Chaotic Crow Search Algorithm, CCSA |
| B61 | 2007 | Bacterial-GA Foraging, BF | B317 | 2018 | Circular Structures of Puffer Fish Algorithm, CSPF |
| B62 | 2007 | Parliamentary Optimization Algorithm, POA | B318 | 2018 | Competitive Learning Algorithm, CLA |
| B63 | 2007 | Simplex Algorithm, SA | B319 | 2018 | Cricket Chirping Algorithm, CCA |
| B64 | 2007 | Good Lattice Swarm Algorithm, GLSA | B320 | 2018 | Fibonacci Indicator Algorithm, FLA |
| B65 | 2007 | Central Force Optimization, CFO | B321 | 2018 | Plant Self-Defense Mechanism Algorithm, PSDM |
| B66 | 2008 | Fast Bacterial Swarming Algorithm, FBSA | B322 | 2018 | Emperor Penguin Optimizer, EPO |
| B67 | 2008 | Biogeography-based Optimization, BBO | B323 | 2018 | Lion Pride Optimization Algorithm, LPOA |
| B68 | 2008 | Bar Systems, BS | B324 | 2018 | Multi-Scale Quantum Harmonic Oscillator Algorithm, MQHO |
| B69 | 2008 | Catfish Particle Swarm Optimization, CatfishPSO | B325 | 2018 | Mushroom Reproduction Optimization, MRO |
| B70 | 2008 | Goose Team Optimizer, GTO | B326 | 2018 | Tree Growth Algorithm, TGA |
| B71 | 2008 | Harmony Element Algorithm, HEA | B327 | 2018 | Moth Search Algorithm, MSA |
| B72 | 2008 | Fish-School Search, FSF | B328 | 2018 | Farmland Fertility, FF |
| B73 | 2008 | Roach Infestation Optimization, RIO | B329 | 2018 | Pity Beetle Algorithm, PBA |
| B74 | 2008 | Viral Search, VS | B330 | 2018 | Mouth Brooding Fish Algorithm, MBF |
| B75 | 2008 | Plant Growth Optimization, PGO | B331 | 2018 | Artificial Flora Optimization Algorithm, AFOA |
| B76 | 2009 | Artificial Beehive Algorithm, ABA | B332 | 2018 | Elephant Swarm Water Search Algorithm, ESWS |
| B77 | 2009 | Artificial Physics Optimization, APO | B333 | 2018 | Sperm Swarm Optimization Algorithm, SSOA |
| B78 | 2009 | Bee Colony-inspired Algorithm, BCiA | B334 | 2018 | Team Game Algorithm, TGA |
| B79 | 2009 | Gravitational Emulation Local Search, GELS | B335 | 2018 | Coyote Optimization Algorithm, COA |
| B80 | 2009 | Group Search Optimizer, GBO | B336 | 2018 | Queuing Search Algorithm, QSA |
| B81 | 2009 | Cuckoo Search, CS | B337 | 2018 | Supernova Optimizer, SO |
| B82 | 2009 | Gravitational Search Algorithm, GSA | B338 | 2018 | Spiritual Search, SS |
| B83 | 2009 | Firefly Algorithm, FA | B339 | 2018 | School Based Optimization, SBO |
| B84 | 2009 | Frog Call inspired Algorithm, FCA | B340 | 2018 | Weighted Vertices Optimizer, WVO |
| B85 | 2009 | Glowworm Swarm Optimization, GSO | B341 | 2018 | Volleyball Premier League Algorithm, VPLA |
| B86 | 2009 | League Championship Algorithm, LCA | B342 | 2018 | Yellow Saddle Goatfish Algorithm, YSGA |
| B87 | 2009 | Paddy Field Algorithm, PFA | B343 | 2019 | Raccoon Optimization Algorithm, ROA |
| B88 | 2009 | Dolphin Partner Optimization, DPO | B344 | 2019 | Andean Condor Algorithm, ACA |
| B89 | 2009 | Dialectic Search, DS | B345 | 2019 | Anglerfish Algorithm, AA |
| B90 | 2009 | Human-Inspired Algorithms, HIA | B346 | 2019 | Artificial Ecosystem-Based Optimization, AEO |
| B91 | 2009 | Artificial Searching Swarm Algorithm, ASSA | B347 | 2019 | Atom Search Optimization Algorithm, ASOA |
| B92 | 2009 | Bumble Bees Mating Optimization, BBMO | B348 | 2019 | Artificial Feeding Birds, AFB |
| B93 | 2009 | Group Counseling Optimization, GCO | B349 | 2019 | Artificial Coronary Circulation System, ACCS |
| B94 | 2009 | Hunting Search Algorithm, HSA | B350 | 2019 | Artificial Electric Field Algorithm, AEFA |
| B95 | 2009 | Locust Swarm, LS | B351 | 2019 | Bus Transportation Algorithm, BTA |
| B96 | 2009 | Intelligent Water Drops Algorithm, IWDA | B352 | 2019 | Biology Migration Algorithm, BMA |
| B97 | 2009 | Water Flow Algorithm, WFA | B353 | 2019 | Buzzard Optimization Algorithm, BUZOA |
| B98 | 2010 | Asexual Reproduction Optimization, ARO | B354 | 2019 | Blue Monkey Algorithm, BM |
| B99 | 2010 | Bean Optimization Algorithm, BOA | B355 | 2019 | Chaotic Dragonfly Algorithm , CDA |
| B100 | 2010 | Bat Algorithm, BA | B356 | 2019 | Cultural Coyote Optimization Algorithm, CCOA |
| B101 | 2010 | Bee Swarm Optimization, BSO | B357 | 2019 | Dice Game Optimizer, DGO |

| Ref | Year | Full name & abbreviation | Ref | Year | Full name & abbreviation |
|---|---|---|---|---|---|
| B102 | 2010 | Charged System Search, CSS | B358 | 2019 | Donkey Theorem Optimization, DTO |
| B103 | 2010 | Chemical Reaction Optimization Algorithm, CRO | B359 | 2019 | Deer Hunting Optimization Algorithm, DHOA |
| B104 | 2010 | Gravitational Field Algorithm, GFA | B360 | 2019 | Falcon Optimization Algorithm, FOA |
| B105 | 2010 | Fireworks Algorithm, FA | B361 | 2019 | Find-Fix-Finish-Exploit-Analyze Algorithm, F3EA |
| B106 | 2010 | Eagle Strategy, ES | B362 | 2019 | Flow Regime Algorithm, FRA |
| B107 | 2010 | Grenade Explosion Algorithm, GEA | B363 | 2019 | Chaotic Optimal Foraging Algorithm, COFA |
| B108 | 2010 | Wind Driven Optimization, WDO | B364 | 2019 | Naked Moled Rat, NMR |
| B109 | 2010 | Termite Colony Optimization, TCO | B365 | 2019 | Xerus Optimization Algorithm, XOA |
| B110 | 2010 | Consultant-Guided Search, CGS | B366 | 2019 | Nuclear Reaction Optimization, NRO |
| B111 | 2010 | Social Emotional Optimization Algorithm, SEOA | B367 | 2019 | Hypercube Natural Aggregation Algorithm, HNAA |
| B112 | 2010 | Hierarchical Swarm Model, HSM | B368 | 2019 | Sailfish Optimizer, SO |
| B113 | 2010 | Reincarnation Algorithm, RA | B369 | 2019 | The Algorithm of the Innovative Gunner, AIG |
| B114 | 2011 | Artificial Plants Optimization Algorithm, APO | B370 | 2019 | Supply-Demand-Based Optimization, SDBO |
| B115 | 2011 | Brain Storm Optimization, BSO | B371 | 2019 | Butterfly Optimization Algorithm, BOA |
| B116 | 2011 | Bioluminescent Swarm Optimization Algorithm, BSOA | B372 | 2019 | Emperor Penguins Colony, EPC |
| B117 | 2011 | Cockroach Swarm Optimization, CSO | B373 | 2019 | Electron Radar Search Algorithm, ERSA |
| B118 | 2011 | Group Escape Behavior, GEB | B374 | 2019 | Henry Gas Solubility Optimization, HGSO |
| B119 | 2011 | Group Leaders Optimization Algorithm, GIOA | B375 | 2019 | Hitchcock Bird-Inspired Algorithm, HBIA |
| B120 | 2011 | Teaching-Learning Base Optimization, TLBO | B376 | 2019 | Hammerhead Shark Optimization Algorithm, HOA |
| B121 | 2011 | Cuckoo Optimization Algorithm, COA | B377 | 2019 | Fitness Dependent Optimizer, FDO |
| B122 | 2011 | Artificial Chemical Reaction Optimization Algorithm, ACROA | B378 | 2019 | Life Choice-Based Optimizer, LCBO |
| B123 | 2011 | Galaxy-Based Search Algorithm, GBSA | B379 | 2019 | Parasitism–Predation Algorithm, PPA |
| B124 | 2011 | Spiral Dynamics Inspired Optimization, SDIO | B380 | 2019 | Pathfinder Algorithm, PA |
| B125 | 2011 | Plant Propagation Algorithm, PPA | B381 | 2019 | Poor And Rich Optimization Algorithm, PROA |
| B126 | 2011 | Eco-Inspired Evolutionary Algorithm, EIEA | B382 | 2019 | Seagull Optimization Algorithm, SOA |
| B127 | 2011 | Gravitational Interactions Optimization, GIO | B383 | 2019 | Sooty Tern Optimization Algorithm, STOA |
| B128 | 2011 | Stem Cells Algorithm, SCA | B384 | 2019 | Harris Hawks Optimization, HHO |
| B129 | 2011 | Water-Flow Algorithm, WFA | B385 | 2019 | Bonobo Optimizer, BO |
| B130 | 2012 | Anarchic Society Optimization, ASO | B386 | 2019 | Spherical Search Optimizer, SSO |
| B131 | 2012 | Artificial Tribe Algorithm, ATA | B387 | 2019 | Squirrel Search Algorithm, SSA |
| B132 | 2012 | Bat Intelligence, BI | B388 | 2019 | Flying Squirrel Optimizer, FSO |
| B133 | 2012 | Collective Animal Behavior, CAB | B389 | 2019 | Bald Eagle Search Optimisation Algorithm, BESO |
| B134 | 2012 | Cloud Model-based Differential Evolution Algorithm, CMDE | B390 | 2019 | Search And Rescue Optimization Algorithm, SAR |
| B135 | 2012 | Flower Pollination Algorithm, FPA | B391 | 2019 | Wild Mice Colony Algorithm, WMC |
| B136 | 2012 | Flock by Leader, FL | B392 | 2019 | Thieves And Police Algorithm, TPA |
| B137 | 2012 | Krill Herd Algorithm, KHA | B393 | 2020 | Artificial Transgender Longicorn Algorithm, ATLA |
| B138 | 2012 | Fruit Fly Optimization Algorithm, FFOA | B394 | 2020 | Barnacles Mating Optimizer, BMO |
| B139 | 2012 | Water Cycle Algorithm, WCA | B395 | 2020 | Black Hole Mechanics Optimization, BHMO |
| B140 | 2012 | Differential Search Algorithm, DSA | B396 | 2020 | Billiards-Inspired Optimization Algorithm, BIOA |
| B141 | 2012 | Ray Optimization, RO | B397 | 2020 | Border Collie Optimization, BCO |
| B142 | 2012 | Migrating Bird Optimization, MBO | B398 | 2020 | Bear Smell Search Algorithm, BSSA |
| B143 | 2012 | Wolf Search Algorithm, WSA | B399 | 2020 | Buyer Inspired Meta-Heuristic Optimization Algorithm, BIMHO |
| B144 | 2012 | Mine Blast Algorithm, MBA | B400 | 2020 | Darts Game Optimizer, DGO |
| B145 | 2012 | Electro-Magnetism Optimization, EMO | B401 | 2020 | Dynamic Differential Annealed Optimization, DDAO |
| B146 | 2012 | Bacterial Colony Optimization, BCO | B402 | 2020 | Dynastic Optimization Algorithm, DOA |
| B147 | 2012 | Great Salmon Run, GSR | B403 | 2020 | Forensic Based Investigation, FBI |
| B148 | 2012 | Japanese Tree Frogs Calling Algorithm, JTFC | B404 | 2020 | Plasma Generation Optimization , PGO |
| B149 | 2012 | Community of Scientist Optimization, CSO | B405 | 2020 | Newton Metaheuristic Algorithm, NMA |
| B150 | 2012 | Quantum-inspired Bacterial Swarming Optimization, QBSO | B406 | 2020 | Tunicate Swarm Algorithm, TSA |
| B151 | 2012 | Hoopoe Heuristic Optimization, HH | B407 | 2020 | Marine Predators Algorithm, MPA |

| Ref | Year | Full name & abbreviation | Ref | Year | Full name & abbreviation |
|---|---|---|---|---|---|
| B152 | 2012 | Intelligent Gravitational Search Algorithm, IGSA | B408 | 2020 | Equilibrium Optimizer, EO |
| B153 | 2012 | Lion Pride Optimizer, LPO | B409 | 2020 | Electric Fish Optimization, EFO |
| B154 | 2012 | Zombie Survival Optimization, ZSO | B410 | 2020 | Slime Mould Algorithm, SMA |
| B155 | 2012 | Artificial Photosynthesis and Phototropism Mechanism, APPM | B411 | 2020 | Black Widow Optimization Algorithm, BWOA |
| B156 | 2012 | Superbug Algorithm, SA | B412 | 2020 | Manta Ray Foraging Optimization, MRFO |
| B157 | 2013 | Artificial Plant Optimization Algorithm, APOA | B413 | 2020 | Mayfly Algorithm, MA |
| B158 | 2013 | Artificial Reaction Algorithm, ARA | B414 | 2020 | Orcas Algorithm, OA |
| B159 | 2013 | Adaptive Social Behavior Optimization, ASBO | B415 | 2020 | Political Optimizer, PO |
| B160 | 2013 | Bat-Inspired Algorithm, BI | B416 | 2020 | Group Teaching Optimization Algorithm, GTOA |
| B161 | 2013 | Co-Operation Of Biology Related Algorithm, COBRA | B417 | 2020 | Turbulent Flow Of Water-Based Optimization, TFWO |
| B162 | 2013 | Global Neighborhood Algorithm, GNA | B418 | 2020 | Human Urbanization Algorithm, HUA |
| B163 | 2013 | Mosquito Host-Seeking Algorithm, MHSA | B419 | 2020 | Chimp Optimization Algorithm, COA |
| B164 | 2013 | Mobility Aware-Termite, MAT | B420 | 2020 | Coronavirus Optimization Algorithm, COA |
| B165 | 2013 | Backtracking Search Optimization, BSO | B421 | 2020 | COVID-19 Optimizer Algorithm, CVA |
| B166 | 2013 | Black Holes Algorithm, BHA | B422 | 2020 | Multivariable Grey Prediction Evolution Algorithm, MGPE |
| B167 | 2013 | Social Spider Optimization, SSO | B423 | 2020 | Sandpiper Optimization Algorithm, SOA |
| B168 | 2013 | Dolphin Echolocation, DE | B424 | 2020 | Shuffled Shepherd Optimization Method, SSOM |
| B169 | 2013 | Artificial Cooperative Search, ACS | B425 | 2020 | Red Deer Algorithm, RDA |
| B170 | 2013 | Gases Brownian Motion Optimization, GBMO | B426 | 2020 | Golden Ratio Optimization Method, GTOM |
| B171 | 2013 | Swallow Swarm Optimization Algorithm, SSOA | B427 | 2020 | Gaining-Sharing Knowledge Based Algorithm, GSKA |
| B172 | 2013 | Penguins Search Optimization Algorithm, PSOA | B428 | 2020 | Adolescent Identity Search Algorithm, AISA |
| B173 | 2013 | Egyptian Vulture Optimization, EVO | B429 | 2020 | Capuchin Search Algorithm, CSA |
| B174 | 2013 | Atmosphere Clouds Model Optimization, ACMO | B430 | 2020 | Giza Pyramids Construction, GPC |
| B175 | 2013 | Magnetotactic Bacteria Optimization Algorithm, MBOA | B431 | 2020 | Grand Tour Algorithm, GTA |
| B176 | 2013 | Blind, Naked Mole-Rats Algorithm, BNMR | B432 | 2020 | Groundwater Flow Algorithm, GFA |
| B177 | 2013 | Soccer Game Optimization, SGO | B433 | 2020 | Gradient-Based Optimizer, GO |
| B178 | 2013 | Seven-Spot Ladybird Optimization, SSLO | B434 | 2020 | Interactive Autodidactic School, IAS |
| B179 | 2013 | Cuttlefish Algorithm, CA | B435 | 2020 | LÉVy Flight Distribution, LFD |
| B180 | 2013 | African Wild Dog Algorithm, AWDA | B436 | 2020 | Momentum Search Algorithm, MSA |
| B181 | 2013 | Mussels Wandering Optimization, MWO | B437 | 2020 | Nomadic People Optimizer, NPO |
| B182 | 2013 | Swine Influenza Models Based Optimization, SIMB | B438 | 2020 | New Caledonian Crow Learning Algorithm, NCCL |
| B183 | 2013 | Tree Physiology Optimization, TPO | B439 | 2020 | Horse Optimization Algorithm, HOA |
| B184 | 2014 | Animal Behavior Hunting, ABH | B440 | 2020 | Rao Algorithms, RA |
| B185 | 2014 | Artificial Raindrop Algorithm, ARA | B441 | 2020 | Rat Swarm Optimizer, RSO |
| B186 | 2014 | Grey Wolf Optimizer, GWO | B442 | 2020 | Rain Optimization Algorithm, ROA |
| B187 | 2014 | Symbiotic Organisms Search, SOS | B443 | 2020 | Student Psychology Based Optimization Algorithm, SPOA |
| B188 | 2014 | Colliding Bodies Optimization, CBO | B444 | 2020 | Seasons Optimization Algorithm, SOA |
| B189 | 2014 | Chicken Swarm Optimization, CSO | B445 | 2020 | Shell Game Optimization, SGO |
| B190 | 2014 | Spider Monkey Optimization, SMO | B446 | 2020 | Sparrow Search Algorithm, SSA |
| B191 | 2014 | Interior Search Algorithm, ISA | B447 | 2020 | Tiki-Taka Algorithm, TTA |
| B192 | 2014 | Animal Migration Optimization Algorithm, AMOA | B448 | 2020 | Transient Search Optimization, TSO |
| B193 | 2014 | Coral Reefs Optimization Algorithm, CROA | B449 | 2020 | Vapor-Liquid Equilibrium Algorithm, VLEA |
| B194 | 2014 | Bird Mating Optimizer, BMO | B450 | 2020 | Virus Spread Optimization, VSO |
| B195 | 2014 | Shark Smell Optimization, SSO | B451 | 2020 | Wingsuit Flying Search, WFS |
| B196 | 2014 | Exchange Market Algorithm, EMA | B452 | 2020 | Water Strider Algorithm, WSA |
| B197 | 2014 | Forest Optimization Algorithm, FOA | B453 | 2020 | Woodpecker Mating Algorithm, WMA |
| B198 | 2014 | Golden Ball Algorithm, GBA | B454 | 2020 | Solar System Algorithm, SSA |
| B199 | 2014 | Keshtel Algorithm, KA | B455 | 2020 | Arsh-Fati-Based Cluster Head Selection Algorithm, ARSH-FATI-CHS |
| B200 | 2014 | Kaizen Programming, KP | B456 | 2020 | Teng-Yue Algorithm, TYA |

| Ref | Year | Full name & abbreviation | Ref | Year | Full name & abbreviation |
|---|---|---|---|---|---|
| B201 | 2014 | Kinetic Gas Molecule Optimization, KGMO | B457 | 2020 | Projectiles Optimization, PO |
| B202 | 2014 | Strawberry Algorithm, SA | B458 | 2020 | Color Harmony Algorithm, CHA |
| B203 | 2014 | Heart Algorithm, HA | B459 | 2020 | Multi-Objective Beetle Antennae Search , MOBAS |
| B204 | 2014 | Artificial Ecosystem Algorithm, AEA | B460 | 2020 | Orca Optimization Algorithm, OOA |
| B205 | 2014 | The Scientific Algorithms, SA | B461 | 2020 | Photon Search Algorithm, PSA |
| B206 | 2014 | Worm Optimization, WO | B462 | 2020 | Kernel Search Optimization, KSO |
| B207 | 2014 | Greedy Politics Optimization, GPO | B463 | 2020 | Spherical Search Algorithm, SSA |
| B208 | 2014 | Human Learning Optimization, HLO | B464 | 2020 | Triple Distinct Search Dynamics, TDSD |
| B209 | 2014 | Soccer League Competition Algorithm, SLCA | B465 | 2021 | Chaos Game Optimization, CGO |
| B210 | 2014 | Hyper-Spherical Search Algorithm, HSSA | B466 | 2021 | Chameleon Swarm Algorithm, CSA |
| B211 | 2014 | Ecogeography-Based Optimization, EBO | B467 | 2021 | Atomic Orbital Search, AOS |
| B212 | 2014 | Pigeon-Inspired Optimization, PIO | B468 | 2021 | Artificial Jellyfish Search Optimizer, JS |
| B213 | 2015 | Ant Lion Optimization, ALO | B469 | 2021 | Cooperation Search Algorithm, CSA |
| B214 | 2015 | Artificial Algae Algorithm, AAA | B470 | 2021 | Material Generation Algorithm, MGA |
| B215 | 2015 | Artificial Showering Algorithm, ASA | B471 | 2021 | Crystal Structure Algorithm, CryStA1 |
| B216 | 2015 | Cricket Algorithm, CA | B472 | 2021 | Archimedes Optimization Algorithm, AOA |
| B217 | 2015 | Gradient Evolution Algorithm, GEA | B473 | 2021 | Archerfish Hunting Optimizer, AHO |
| B218 | 2015 | Moth-Flame Optimization Algorithm, MFOA | B474 | 2021 | Battle Royale Optimization Algorithm, BRO |
| B219 | 2015 | Monarch Butterfly Optimization, MBO | B475 | 2021 | Artificial Lizard Search Optimization, ALSO |
| B220 | 2015 | Water Wave Optimization, WWO | B476 | 2021 | Quantum Firefly Algorithm, QFA |
| B221 | 2015 | Stochastic Fractal Search, SFS | B477 | 2021 | Flow Direction Algorithm, FDA |
| B222 | 2015 | Elephant Herding Optimization, EHO | B478 | 2021 | Lichtenberg Algorithm, LA |
| B223 | 2015 | Vortex Search Algorithm, VSA | B479 | 2021 | Pastoralist Optimization Algorithm, POA |
| B224 | 2015 | Earthworm Optimization Algorithm, EOA | B480 | 2021 | Ebola Optimization Search Algorithm, EOSA |
| B225 | 2015 | Lightning Search Algorithm, LSA | B481 | 2021 | Elephant Clan Optimization, ECO |
| B226 | 2015 | Heat Transfer Search Algorithm, HTSA | B482 | 2021 | Red Colobuses Monkey, RCM |
| B227 | 2015 | Ions Motion Algorithm, IMA | B483 | 2021 | Golden Eagle Optimizer, GEO |
| B228 | 2015 | Optics Inspired Optimization, OIO | B484 | 2021 | Group Mean-Based Optimizer, GMBO |
| B229 | 2015 | Tree Seed Algorithm, TSA | B485 | 2021 | Dingo Optimizer, DO |
| B230 | 2015 | Runner-Root Algorithm, RRA | B486 | 2021 | Coronavirus Herd Immunity Optimizer, CHIO |
| B231 | 2015 | Elephant Search Algorithm, ESA | B487 | 2021 | Red Fox Optimization Algorithm, RFO |
| B232 | 2015 | Election Algorithm, EA | B488 | 2021 | Arithmetic Optimization Algorithm, AOA |
| B233 | 2015 | Locust Search, LS | B489 | 2021 | African Vultures Optimization Algorithm, AVOA |
| B234 | 2015 | Invasive Tumor Growth Optimization Algorithm, ITWO | B490 | 2021 | Artificial Gorilla Troops Optimizer, GTO |
| B235 | 2015 | Jaguar Algorithm, JA | B491 | 2021 | Artificial Hummingbird Algorithm, AHA |
| B236 | 2015 | General Relativity Search Algorithm, GRSA | B492 | 2021 | Intelligent Ice Fishing Algorithm, IIFA |
| B237 | 2015 | Root Growth Optimizer, RGO | B493 | 2021 | Komodo Mlipir Algorithm, KMA |
| B238 | 2015 | Bull Optimization Algorithm, BOA | B494 | 2021 | Linear Prediction Evolution Algorithm, LPE |
| B239 | 2015 | Prey-Predator Algorithm, PPA | B495 | 2021 | Multi-Objective Trader Algorithm, MOTR |
| B240 | 2015 | African Buffalo Optimization, ABO | B496 | 2021 | Optimal Stochastic Process Optimizer, OSPO |
| B241 | 2016 | Artificial Infectious Disease Optimization, AID | B497 | 2021 | Remora Optimization Algorithm, ROA |
| B242 | 2016 | Across Neighborhood Search, ANS | B498 | 2021 | Ring Toss Game-Based Optimization Algorithm, RTGBO |
| B243 | 2016 | Cricket Behavior-Based Algorithm, CBBA | B499 | 2021 | RUNge Kutta Optimizer, RUN |
| B244 | 2016 | Competitive Optimization Algorithm, COOA | B500 | 2021 | Samw |
| B245 | 2016 | Cognitive Behavior Optimization Algorithm, COA | B501 | 2021 | String Theory Algorithm, STA |
| B246 | 2016 | Electromagnetic Field Optimization, EFO | B502 | 2021 | Success History Intelligent Optimizer, SHIO |
| B247 | 2016 | Football Game Algorithm, FGA | B503 | 2021 | Tangent Search Algorithm, TSA |
| B248 | 2016 | Intrusive Tumor Growth Inspired Optimization Algorithm, ITGO | B504 | 2021 | Tuna Swarm Optimization, TSO |
| B249 | 2016 | Galactic Swarm Optimization, GSO | B505 | 2021 | Volcano Eruption Algorithm, VCA |
| B250 | 2016 | Whale Optimization Algorithm, WOA | B506 | 2021 | Smart Flower Optimization Algorithm, SFOA |
| B251 | 2016 | Sine Cosine Algorithm, SSA | B507 | 2022 | Ali baba and the Forty Thieves Optimization, AFT |

| Ref | Year | Full name & abbreviation | Ref | Year | Full name & abbreviation |
|---|---|---|---|---|---|
| B252 | 2016 | Dragonfly Algorithm, DA | B508 | 2022 | Honey Badger Algorithm, HBA |
| B253 | 2016 | Crow Search Algorithm, CSA | B509 | 2022 | Orca Predation Algorithm, OPA |
| B254 | 2016 | Multi-Verse Optimizer, MVO | B510 | 2022 | Reptile Search Algorithm, RSA |
| B255 | 2016 | Bird Swarm Algorithm, BSA | B511 | 2022 | Skip Salp Swam Algorithm, SSSA |
| B256 | 2016 | Virus Colony Search, VCS | | | |

# Reference

[1] G. Wu, R. Mallipeddi, P.N. Suganthan, Ensemble strategies for population-based optimization algorithms – A survey, Swarm Evol. Comput. 44 (2019) 695–711. https://doi.org/10.1016/j.swevo.2018.08.015.

[2] R. Moghdani, K. Salimifard, Volleyball premier league algorithm, Appl. Soft Comput. 64 (2018) 161–185.

[3] R. Rajabioun, Cuckoo Optimization Algorithm, Appl. Soft Comput. 11 (2011) 5508–5518. https://doi.org/10.1016/j.asoc.2011.05.008.

[4] F. Glover, Future paths for integer programming and links to artificial intelligence, Comput. Oper. Res. 13 (1986) 533–549.

[5] Q. Askari, I. Younas, M. Saeed, Political Optimizer: A novel socio-inspired meta-heuristic for global optimization, Knowledge-Based Syst. 195 (2020). https://doi.org/10.1016/j.knosys.2020.105709.

[6] S. Kirkpatrick, C.D. Gelatt, M.P. Vecchi, Optimization by simulated annealing, Science (80-. ). 220 (1983) 671–680.

[7] J. Kennedy, R. Eberhart, Particle swarm optimization, in: Proc. ICNN'95-International Conf. Neural Networks, IEEE, 1995: pp. 1942–1948.

[8] M. Dorigo, M. Birattari, T. Stutzle, Ant colony optimization, IEEE Comput. Intell. Mag. 1 (2006) 28–39.

[9] S.H. Aghay Kaboli, J. Selvaraj, N.A. Rahim, Rain-fall optimization algorithm: A population based algorithm for solving constrained optimization problems, J. Comput. Sci. 19 (2017) 31–42. https://doi.org/10.1016/j.jocs.2016.12.010.

[10] H.R. Lourenço, O.C. Martin, T. Stützle, Iterated local search: Framework and applications, in: Handb. Metaheuristics, Springer, 2019: pp. 129–168.

[11] C. Voudouris, E.P.K. Tsang, A. Alsheddy, Guided local search, in: Handb. Metaheuristics, Springer, 2010: pp. 321–361.

[12] L.A. Rastrigin, The convergence of the random search method in the extremal control of a many parameter system, Autom. Remote Control. 24 (1963) 1337–1342.

[13] N. Mladenović, P. Hansen, Variable neighborhood search, Comput. Oper. Res. 24 (1997) 1097–1100.

[14] D. Pisinger, S. Ropke, Large neighborhood search, in: Handb. Metaheuristics, Springer, 2010: pp. 399–419.

[15] J.H. Holland, Adaptation in natural and artificial systems, univ. of mich. press, Ann Arbor. (1975).

[16] R. Storn, K. Price, Differential evolution–a simple and efficient heuristic for global optimization over continuous spaces, J. Glob. Optim. 11 (1997) 341–359.

[17] R. Hooke, T.A. Jeeves, ``Direct Search''Solution of Numerical and Statistical Problems, J. ACM. 8 (1961) 212–229.

[18] A. Tzanetos, G. Dounias, Nature inspired optimization algorithms or simply variations of metaheuristics?, Artif. Intell. Rev. 54 (2021) 1841–1862.

[19] D. Molina, J. Poyatos, J. Del Ser, S. García, A. Hussain, F. Herrera, Comprehensive Taxonomies of Nature- and Bio-inspired Optimization: Inspiration Versus Algorithmic Behavior, Critical Analysis Recommendations, Cognit. Comput. 12 (2020) 897–939. https://doi.org/10.1007/s12559-020-09730-8.

[20] A.H. Halim, I. Ismail, S. Das, Performance assessment of the metaheuristic optimization algorithms: an exhaustive review, Artif. Intell. Rev. (2020). https://doi.org/10.1007/s10462-020-09906-6.

[21] X. Chu, T. Wu, J.D. Weir, Y. Shi, B. Niu, L. Li, Learning–interaction–diversification framework for swarm intelligence optimizers: a unified perspective, Neural Comput. Appl. 32 (2018) 1789–1809. https://doi.org/10.1007/s00521-018-3657-0.

[22] Z. Chen, Y. Liu, Z. Yang, X. Fu, J. Tan, X. Yang, An enhanced teaching-learning-based optimization algorithm with self-adaptive and learning operators and its search bias towards origin, Swarm Evol. Comput. 60 (2021). https://doi.org/10.1016/j.swevo.2020.100766.

[23] J.K. Pickard, J.A. Carretero, V.C. Bhavsar, On the convergence and origin bias of the Teaching-Learning-Based-Optimization algorithm, Appl. Soft Comput. 46 (2016) 115–127. https://doi.org/10.1016/j.asoc.2016.04.029.

[24] P. Niu, S. Niu, N. liu, L. Chang, The defect of the Grey Wolf optimization algorithm and its verification method, Knowledge-Based Syst. 171 (2019) 37–43. https://doi.org/https://doi.org/10.1016/j.knosys.2019.01.018.

[25] A.E. Ezugwu, A.K. Shukla, R. Nath, A.A. Akinyelu, J.O. Agushaka, H. Chiroma, P.K. Muhuri, Metaheuristics: a comprehensive overview and classification along with bibliometric analysis, Artif. Intell. Rev. (2021). https://doi.org/10.1007/s10462-020-09952-0.

[26] C. Pizzuti, Evolutionary computation for community detection in networks: a review, IEEE Trans. Evol. Comput. 22 (2017) 464–483.

[27] N.A. Barricelli, Symbiogenetic evolution processes realized by artificial methods, 1957.

[28] A.S. Fraser, Simulation of genetic systems by automatic digital computers I. Introduction, Aust. J. Biol. Sci. 10 (1957) 484–491.

[29] I. Rechenberg, Evolution strategy: Nature’s way of optimization, in: Optim. Methods Appl. Possibilities Limitations, Springer, 1989: pp. 106–126.

[30] L.J. Fogel, A.J. Owens, M.J. Walsh, Intelligent decision making through a simulation of evolution, Behav. Sci. 11 (1966) 253–272.

[31] D. Simon, Evolutionary optimization algorithms, John Wiley & Sons, 2013.

[32] D.B. Fogel, Evolutionary computation: toward a new philosophy of machine intelligence, John Wiley & Sons, 2006.

[33] L.J. Fogel, Artificial intelligence through a simulation of evolution, in: Proc. 2nd Cybern. Sci. Symp., 1965, 1965.

[34] X. Yao, Y. Liu, G. Lin, Evolutionary programming made faster, IEEE Trans. Evol. Comput. 3 (1999) 82–102.

[35] R. Mallipeddi, P.N. Suganthan, Evaluation of novel adaptive evolutionary programming on four constraint handling techniques, in: 2008 IEEE Congr. Evol. Comput. (IEEE World Congr. Comput. Intell., IEEE, 2008: pp. 4045–4052.

[36] H. Zhang, J. Lu, Adaptive evolutionary programming based on reinforcement learning, Inf. Sci. (Ny). 178 (2008) 971–984.

[37] R. Mallipeddi, S. Mallipeddi, P.N. Suganthan, Ensemble strategies with adaptive evolutionary programming, Inf. Sci. (Ny). 180 (2010) 1571–1581.

[38] Q. Liu, X. Li, L. Gao, Y. Li, A modified genetic algorithm with new encoding and decoding methods for integrated process planning and scheduling problem, IEEE Trans. Cybern. (2020).

[39] G. Zhang, Y. Hu, J. Sun, W. Zhang, An improved genetic algorithm for the flexible job shop scheduling problem with multiple time constraints, Swarm Evol. Comput. 54 (2020) 100664.

[40] S. Sayed, M. Nassef, A. Badr, I. Farag, A nested genetic algorithm for feature selection in high-dimensional cancer microarray datasets, Expert Syst. Appl. 121 (2019) 233–243.

[41] F. Curtis, X. Li, T. Rose, A. Vazquez-Mayagoitia, S. Bhattacharya, L.M. Ghiringhelli, N. Marom, GAtor: a first-principles genetic algorithm for molecular crystal structure prediction, J. Chem. Theory Comput. 14 (2018) 2246–2264.

[42] W. Chen, M. Panahi, H.R. Pourghasemi, Performance evaluation of GIS-based new ensemble data mining techniques of adaptive neuro-fuzzy inference system (ANFIS) with genetic algorithm (GA), differential evolution (DE), and particle swarm optimization (PSO) for landslide spatial modelling, Catena. 157 (2017) 310–324.

[43] E.C. Pedrino, T. Yamada, T.R. Lunardi, J.C. de Melo Vieira Jr, Islanding detection of distributed generation by using multi-gene genetic programming based classifier, Appl. Soft Comput. 74 (2019) 206–215.

[44] C. Ryan, J.J. Collins, M.O. Neill, Grammatical evolution: Evolving programs for an arbitrary language, in: Eur. Conf. Genet. Program., Springer, 1998: pp. 83–96.

[45] S. Das, P.N. Suganthan, Differential evolution: A survey of the state-of-the-art, IEEE Trans. Evol. Comput. 15 (2010) 4–31.

[46] G. Wu, R. Mallipeddi, P.N. Suganthan, R. Wang, H. Chen, Differential evolution with multi-population based ensemble of mutation strategies, Inf. Sci. (Ny). 329 (2016) 329–345.

[47] G. Wu, X. Shen, H. Li, H. Chen, A. Lin, P.N. Suganthan, Ensemble of differential evolution variants, Inf. Sci. (Ny). 423 (2018) 172–186. https://doi.org/https://doi.org/10.1016/j.ins.2017.09.053.

[48] A.K. Qin, V.L. Huang, P.N. Suganthan, Differential evolution algorithm with strategy adaptation for global numerical optimization, IEEE Trans. Evol. Comput. 13 (2008) 398–417.

[49] J. Brest, S. Greiner, B. Boskovic, M. Mernik, V. Zumer, Self-adapting control parameters in differential evolution: A comparative study on numerical benchmark problems, IEEE Trans. Evol. Comput. 10 (2006) 646–657.

[50] Y. Wang, Z. Cai, Q. Zhang, Differential evolution with composite trial vector generation strategies and control parameters, IEEE Trans. Evol. Comput. 15 (2011) 55–66.

[51] F.A. Hashim, E.H. Houssein, M.S. Mabrouk, W. Al-Atabany, S. Mirjalili, Henry gas solubility optimization: A novel physics-based algorithm, Futur. Gener. Comput. Syst. 101 (2019) 646–667. https://doi.org/10.1016/j.future.2019.07.015.

[52] A.W. Mohamed, A.A. Hadi, A.K. Mohamed, Gaining-sharing knowledge based algorithm for solving optimization problems: a novel nature-inspired algorithm, Int. J. Mach. Learn. Cybern. 11 (2020) 1501–1529. https://doi.org/10.1007/s13042-019-01053-x.

[53] G. Dhiman, M. Garg, A. Nagar, V. Kumar, M. Dehghani, A novel algorithm for global optimization: Rat Swarm Optimizer, J. Ambient Intell. Humaniz. Comput. (2020). https://doi.org/10.1007/s12652-020-02580-0.

[54] W. Al-Sorori, A.M. Mohsen, New Caledonian crow learning algorithm: A new metaheuristic algorithm for solving continuous optimization problems, Appl. Soft Comput. 92 (2020). https://doi.org/10.1016/j.asoc.2020.106325.

[55] M.H. Sulaiman, Z. Mustaffa, M.M. Saari, H. Daniyal, Barnacles Mating Optimizer: A new bio-inspired algorithm for solving engineering optimization problems, Eng. Appl. Artif. Intell. 87 (2020). https://doi.org/10.1016/j.engappai.2019.103330.

[56] R. V Rao, V.J. Savsani, D.P. Vakharia, Teaching–learning-based optimization: A novel method for constrained mechanical design optimization problems, Comput. Des. 43 (2011) 303–315. https://doi.org/https://doi.org/10.1016/j.cad.2010.12.015.

[57] Y. Zhang, Z. Jin, Group teaching optimization algorithm: A novel metaheuristic method for solving global optimization problems, Expert Syst. Appl. 148 (2020). https://doi.org/10.1016/j.eswa.2020.113246.

[58] N. Moosavian, B. Kasaee Roodsari, Soccer league competition algorithm: A novel meta-heuristic algorithm for optimal design of water distribution networks, Swarm Evol. Comput. 17 (2014) 14–24.

https://doi.org/10.1016/j.swevo.2014.02.002.
[59] E. Atashpaz-Gargari, C. Lucas, Imperialist competitive algorithm: An algorithm for optimization inspired by imperialistic competition, in: 2007 IEEE Congr. Evol. Comput., 2007: pp. 4661–4667. https://doi.org/10.1109/CEC.2007.4425083.
[60] J.S.M.L. Melvix, Greedy politics optimization: Metaheuristic inspired by political strategies adopted during state assembly elections, in: 2014 IEEE Int. Adv. Comput. Conf., IEEE, 2014: pp. 1157–1162.
[61] A. Borji, A new global optimization algorithm inspired by parliamentary political competitions, in: Mex. Int. Conf. Artif. Intell., Springer, 2007: pp. 61–71.
[62] S.H. Samareh Moosavi, V.K. Bardsiri, Poor and rich optimization algorithm: A new human-based and multi populations algorithm, Eng. Appl. Artif. Intell. 86 (2019) 165–181. https://doi.org/10.1016/j.engappai.2019.08.025.
[63] H. Ghasemian, F. Ghasemian, H. Vahdat-Nejad, Human urbanization algorithm: A novel metaheuristic approach, Math. Comput. Simul. 178 (2020) 1–15. https://doi.org/10.1016/j.matcom.2020.05.023.
[64] A. Khatri, A. Gaba, K.P.S. Rana, V. Kumar, A novel life choice-based optimizer, Soft Comput. 24 (2019) 9121–9141. https://doi.org/10.1007/s00500-019-04443-z.
[65] J. Zhang, M. Xiao, L. Gao, Q. Pan, Queuing search algorithm: A novel metaheuristic algorithm for solving engineering optimization problems, Appl. Math. Model. 63 (2018) 464–490. https://doi.org/10.1016/j.apm.2018.06.036.
[66] Q. Zhang, R. Wang, J. Yang, A. Lewis, F. Chiclana, S. Yang, Biology migration algorithm: a new nature-inspired heuristic methodology for global optimization, Soft Comput. 23 (2019) 7333–7358.
[67] D. Karaboga, B. Basturk, A powerful and efficient algorithm for numerical function optimization: artificial bee colony (ABC) algorithm, J. Glob. Optim. 39 (2007) 459–471. https://doi.org/10.1007/s10898-007-9149-x.
[68] X.-S. Yang, A new metaheuristic bat-inspired algorithm, in: Nat. Inspired Coop. Strateg. Optim. (NICSO 2010), Springer, 2010: pp. 65–74.
[69] X.-S. Yang, S. Deb, Cuckoo search via Lévy flights, in: 2009 World Congr. Nat. Biol. Inspired Comput., Ieee, 2009: pp. 210–214.
[70] X.-S. Yang, Firefly algorithms for multimodal optimization, in: Int. Symp. Stoch. Algorithms, Springer, 2009: pp. 169–178.
[71] S. Mirjalili, S.M. Mirjalili, A. Lewis, Grey Wolf Optimizer, Adv. Eng. Softw. 69 (2014) 46–61. https://doi.org/10.1016/j.advengsoft.2013.12.007.
[72] S.Z. Mirjalili, S. Mirjalili, S. Saremi, H. Faris, I. Aljarah, Grasshopper optimization algorithm for multi-objective optimization problems, Appl. Intell. 48 (2018) 805–820.
[73] S. Harifi, J. Mohammadzadeh, M. Khalilian, S. Ebrahimnejad, Giza Pyramids Construction: an ancient-inspired metaheuristic algorithm for optimization, Evol. Intell. (2020) 1–19.
[74] A.R. Mehrabian, C. Lucas, A novel numerical optimization algorithm inspired from weed colonization, Ecol. Inform. 1 (2006) 355–366.
[75] X.-S. Yang, Flower pollination algorithm for global optimization, in: Int. Conf. Unconv. Comput. Nat. Comput., Springer, 2012: pp. 240–249.
[76] S. Harifi, M. Khalilian, J. Mohammadzadeh, S. Ebrahimnejad, Emperor Penguins Colony: a new metaheuristic algorithm for optimization, Evol. Intell. 12 (2019) 211–226.
[77] V. Hayyolalam, A.A.P. Kazem, Black widow optimization algorithm: a novel meta-heuristic approach for solving engineering optimization problems, Eng. Appl. Artif. Intell. 87 (2020) 103249.
[78] E. Bogar, S. Beyhan, Adolescent Identity Search Algorithm (AISA): A novel metaheuristic approach for

solving optimization problems, Appl. Soft Comput. 95 (2020). https://doi.org/10.1016/j.asoc.2020.106503.
[79] M. Khishe, M.R. Mosavi, Chimp optimization algorithm, Expert Syst. Appl. 149 (2020) 113338.
[80] E. Rashedi, H. Nezamabadi-pour, S. Saryazdi, GSA: A Gravitational Search Algorithm, Inf. Sci. (Ny). 179 (2009) 2232–2248. https://doi.org/10.1016/j.ins.2009.03.004.
[81] S. Mirjalili, S.M. Mirjalili, A. Hatamlou, Multi-Verse Optimizer: a nature-inspired algorithm for global optimization, Neural Comput. Appl. 27 (2015) 495–513. https://doi.org/10.1007/s00521-015-1870-7.
[82] Z.W. Geem, J.H. Kim, G.V. Loganathan, A new heuristic optimization algorithm: harmony search, Simulation. 76 (2001) 60–68.
[83] T. Zhang, Z.W. Geem, Review of harmony search with respect to algorithm structure, Swarm Evol. Comput. 48 (2019) 31–43. https://doi.org/https://doi.org/10.1016/j.swevo.2019.03.012.
[84] A. Kaveh, T. Bakhshpoori, Water Evaporation Optimization: A novel physically inspired optimization algorithm, Comput. Struct. 167 (2016) 69–85. https://doi.org/10.1016/j.compstruc.2016.01.008.
[85] M.H. Qais, H.M. Hasanien, S. Alghuwainem, Transient search optimization: a new meta-heuristic optimization algorithm, Appl. Intell. 50 (2020) 3926–3941. https://doi.org/10.1007/s10489-020-01727-y.
[86] A.Y.S. Lam, V.O.K. Li, Chemical-reaction-inspired metaheuristic for optimization, IEEE Trans. Evol. Comput. 14 (2009) 381–399.
[87] A. Kaveh, S. Talatahari, A novel heuristic optimization method: charged system search, Acta Mech. 213 (2010) 267–289. https://doi.org/10.1007/s00707-009-0270-4.
[88] E.-G. Talbi, Metaheuristics: from design to implementation, John Wiley & Sons, 2009.
[89] B. Doğan, T. Ölmez, A new metaheuristic for numerical function optimization: Vortex Search algorithm, Inf. Sci. (Ny). 293 (2015) 125–145. https://doi.org/10.1016/j.ins.2014.08.053.
[90] F.A. Hashim, K. Hussain, E.H. Houssein, M.S. Mabrouk, W. Al-Atabany, Archimedes optimization algorithm: a new metaheuristic algorithm for solving optimization problems, Appl. Intell. (2020). https://doi.org/10.1007/s10489-020-01893-z.
[91] A. Faramarzi, M. Heidarinejad, B. Stephens, S. Mirjalili, Equilibrium optimizer: A novel optimization algorithm, Knowledge-Based Syst. 191 (2020) 105190. https://doi.org/https://doi.org/10.1016/j.knosys.2019.105190.
[92] G. Dhiman, V. Kumar, Seagull optimization algorithm: Theory and its applications for large-scale industrial engineering problems, Knowledge-Based Syst. 165 (2019) 169–196. https://doi.org/10.1016/j.knosys.2018.11.024.
[93] B. Morales-Castañeda, D. Zaldívar, E. Cuevas, F. Fausto, A. Rodríguez, A better balance in metaheuristic algorithms: Does it exist?, Swarm Evol. Comput. 54 (2020). https://doi.org/10.1016/j.swevo.2020.100671.
[94] M. Črepinšek, S.-H. Liu, M. Mernik, Exploration and exploitation in evolutionary algorithms: A survey, ACM Comput. Surv. 45 (2013) 1–33.
[95] Z.-H. Zhan, Z.-J. Wang, H. Jin, J. Zhang, Adaptive distributed differential evolution, IEEE Trans. Cybern. 50 (2019) 4633–4647.
[96] K.R. Opara, J. Arabas, Differential Evolution: A survey of theoretical analyses, Swarm Evol. Comput. 44 (2019) 546–558.
[97] A.W. Mohamed, A.A. Hadi, K.M. Jambi, Novel mutation strategy for enhancing SHADE and LSHADE algorithms for global numerical optimization, Swarm Evol. Comput. 50 (2019) 100455. https://doi.org/https://doi.org/10.1016/j.swevo.2018.10.006.
[98] A. Viktorin, R. Senkerik, M. Pluhacek, T. Kadavy, A. Zamuda, Distance based parameter adaptation for Success-History based Differential Evolution, Swarm Evol. Comput. 50 (2019) 100462. https://doi.org/https://doi.org/10.1016/j.swevo.2018.10.013.

[99] N.H. Awad, M.Z. Ali, P.N. Suganthan, Ensemble sinusoidal differential covariance matrix adaptation with Euclidean neighborhood for solving CEC2017 benchmark problems, in: 2017 IEEE Congr. Evol. Comput., 2017: pp. 372–379. https://doi.org/10.1109/CEC.2017.7969336.

[100] A.W. Mohamed, A.A. Hadi, A.M. Fattouh, K.M. Jambi, LSHADE with semi-parameter adaptation hybrid with CMA-ES for solving CEC 2017 benchmark problems, in: 2017 IEEE Congr. Evol. Comput., IEEE, 2017: pp. 145–152.

[101] A. Ghosh, S. Das, A.K. Das, R. Senkerik, A. Viktorin, I. Zelinka, A.D. Masegosa, Using spatial neighborhoods for parameter adaptation: An improved success history based differential evolution, Swarm Evol. Comput. 71 (2022) 101057.

[102] G. Zhang, Y. Shi, Hybrid Sampling Evolution Strategy for Solving Single Objective Bound Constrained Problems, in: 2018 IEEE Congr. Evol. Comput., 2018: pp. 1–7. https://doi.org/10.1109/CEC.2018.8477908.

[103] A. Kumar, R.K. Misra, D. Singh, Improving the local search capability of Effective Butterfly Optimizer using Covariance Matrix Adapted Retreat Phase, in: 2017 IEEE Congr. Evol. Comput., 2017: pp. 1835–1842. https://doi.org/10.1109/CEC.2017.7969524.

[104] H. Rakhshani, A. Rahati, Snap-drift cuckoo search: A novel cuckoo search optimization algorithm, Appl. Soft Comput. 52 (2017) 771–794. https://doi.org/https://doi.org/10.1016/j.asoc.2016.09.048.

[105] X.-S. Yang, S. Deb, Cuckoo search: recent advances and applications, Neural Comput. Appl. 24 (2014) 169–174. https://doi.org/10.1007/s00521-013-1367-1.

[106] H. Chen, M. Wang, X. Zhao, A multi-strategy enhanced sine cosine algorithm for global optimization and constrained practical engineering problems, Appl. Math. Comput. 369 (2020) 124872. https://doi.org/https://doi.org/10.1016/j.amc.2019.124872.

[107] S. Mirjalili, SCA: A Sine Cosine Algorithm for solving optimization problems, Knowledge-Based Syst. 96 (2016) 120–133. https://doi.org/10.1016/j.knosys.2015.12.022.

[108] D. Pelusi, R. Mascella, L. Tallini, J. Nayak, B. Naik, Y. Deng, An Improved Moth-Flame Optimization algorithm with hybrid search phase, Knowledge-Based Syst. 191 (2020) 105277. https://doi.org/https://doi.org/10.1016/j.knosys.2019.105277.

[109] S. Mirjalili, Moth-flame optimization algorithm: A novel nature-inspired heuristic paradigm, Knowledge-Based Syst. 89 (2015) 228–249. https://doi.org/10.1016/j.knosys.2015.07.006.

[110] L. Abualigah, D. Yousri, M. Abd Elaziz, A.A. Ewees, M.A.A. Al-Qaness, A.H. Gandomi, Aquila optimizer: a novel meta-heuristic optimization algorithm, Comput. Ind. Eng. 157 (2021) 107250.

[111] J. Luo, H. Chen, Q. zhang, Y. Xu, H. Huang, X. Zhao, An improved grasshopper optimization algorithm with application to financial stress prediction, Appl. Math. Model. 64 (2018) 654–668. https://doi.org/https://doi.org/10.1016/j.apm.2018.07.044.

[112] D. Pelusi, R. Mascella, L. Tallini, J. Nayak, B. Naik, Y. Deng, Improving exploration and exploitation via a Hyperbolic Gravitational Search Algorithm, Knowledge-Based Syst. 193 (2020) 105404. https://doi.org/https://doi.org/10.1016/j.knosys.2019.105404.

[113] M.M. Eusuff, K.E. Lansey, Optimization of water distribution network design using the shuffled frog leaping algorithm, J. Water Resour. Plan. Manag. 129 (2003) 210–225.

[114] D. Tang, Z. Liu, J. Yang, J. Zhao, Memetic frog leaping algorithm for global optimization, Soft Comput. 23 (2019) 11077–11105. https://doi.org/10.1007/s00500-018-3662-3.

[115] A. Faramarzi, M. Heidarinejad, S. Mirjalili, A.H. Gandomi, Marine Predators Algorithm: A nature-inspired metaheuristic, Expert Syst. Appl. 152 (2020) 113377. https://doi.org/https://doi.org/10.1016/j.eswa.2020.113377.

[116] S. Debnath, W. Arif, S. Baishya, Buyer Inspired Meta-Heuristic Optimization Algorithm, Open Comput. Sci. 10 (2020) 194–219. https://doi.org/10.1515/comp-2020-0101.

[117] A. LaTorre, D. Molina, E. Osaba, J. Poyatos, J. Del Ser, F. Herrera, A prescription of methodological guidelines for comparing bio-inspired optimization algorithms, Swarm Evol. Comput. 67 (2021) 100973.

[118] E. Osaba, E. Villar-Rodriguez, J. Del Ser, A.J. Nebro, D. Molina, A. LaTorre, P.N. Suganthan, C.A.C. Coello, F. Herrera, A tutorial on the design, experimentation and application of metaheuristic algorithms to real-world optimization problems, Swarm Evol. Comput. 64 (2021) 100888.

[119] M. López-Ibáñez, J. Dubois-Lacoste, L. Pérez Cáceres, M. Birattari, T. Stützle, The irace package: Iterated racing for automatic algorithm configuration, Oper. Res. Perspect. 3 (2016) 43–58. https://doi.org/https://doi.org/10.1016/j.orp.2016.09.002.

[120] L.C.T. Bezerra, M. López-Ibáñez, T. Stützle, Automatically Designing State-of-the-Art Multi- and Many-Objective Evolutionary Algorithms, Evol. Comput. 28 (2019) 195–226. https://doi.org/10.1162/evco_a_00263.

[121] A.J. Nebro, M. López-Ibáñez, C. Barba-González, J. García-Nieto, Automatic configuration of NSGA-II with jMetal and irace, Proc. Genet. Evol. Comput. Conf. Companion. (2019) 1374–1381. https://doi.org/10.1145/3319619.3326832.

[122] C. Huang, Y. Li, X. Yao, A Survey of Automatic Parameter Tuning Methods for Metaheuristics, IEEE Trans. Evol. Comput. 24 (2020) 201–216. https://doi.org/10.1109/TEVC.2019.2921598.

[123] D. Jiang, X. Li, Order fulfilment problem with time windows and synchronisation arising in the online retailing, Int. J. Prod. Res. 59 (2021) 1187–1215.

[124] J.J. Liang, S. Baskar, P.N. Suganthan, A.K. Qin, Performance Evaluation of Multiagent Genetic Algorithm, Nat. Comput. 5 (2006) 83–96. https://doi.org/10.1007/s11047-005-1625-y.

[125] J. Carrasco, S. García, M.M. Rueda, S. Das, F. Herrera, Recent trends in the use of statistical tests for comparing swarm and evolutionary computing algorithms: Practical guidelines and a critical review, Swarm Evol. Comput. 54 (2020). https://doi.org/10.1016/j.swevo.2020.100665.

[126] J. Derrac, S. García, D. Molina, F. Herrera, A practical tutorial on the use of nonparametric statistical tests as a methodology for comparing evolutionary and swarm intelligence algorithms, Swarm Evol. Comput. 1 (2011) 3–18.

[127] L.C.T. Bezerra, M. López-Ibánez, T. Stützle, Automatic component-wise design of multiobjective evolutionary algorithms, IEEE Trans. Evol. Comput. 20 (2015) 403–417.

[128] M. López-Ibánez, T. Stützle, Automatically improving the anytime behaviour of optimisation algorithms, Eur. J. Oper. Res. 235 (2014) 569–582.

[129] A.P. Piotrowski, J.J. Napiorkowski, Some metaheuristics should be simplified, Inf. Sci. (Ny). 427 (2018) 32–62.

[130] E.-G. Talbi, Combining metaheuristics with mathematical programming, constraint programming and machine learning, Ann. Oper. Res. 240 (2016) 171–215.

[131] M. Karimi-Mamaghan, M. Mohammadi, P. Meyer, A.M. Karimi-Mamaghan, E.-G. Talbi, Machine learning at the service of meta-heuristics for solving combinatorial optimization problems: A state-of-the-art, Eur. J. Oper. Res. 296 (2022) 393–422.

[132] E.-G. Talbi, Machine Learning into Metaheuristics: A Survey and Taxonomy, ACM Comput. Surv. 54 (2021) 1–32.

[133] J. Ma, D. Xia, Y. Wang, X. Niu, S. Jiang, Z. Liu, H. Guo, A comprehensive comparison among metaheuristics (MHs) for geohazard modeling using machine learning: Insights from a case study of landslide displacement prediction, Eng. Appl. Artif. Intell. 114 (2022) 105150.

[134] S. Nematzadeh, F. Kiani, M. Torkamanian-Afshar, N. Aydin, Tuning hyperparameters of machine learning algorithms and deep neural networks using metaheuristics: A bioinformatics study on biomedical and biological cases, Comput. Biol. Chem. 97 (2022) 107619.
[135] J.-S. Chou, T.-K. Nguyen, Forward forecast of stock price using sliding-window metaheuristic-optimized machine-learning regression, IEEE Trans. Ind. Informatics. 14 (2018) 3132–3142.
[136] A. Song, G. Wu, W. Pedrycz, L. Wang, Integrating Variable Reduction Strategy With Evolutionary Algorithms for Solving Nonlinear Equations Systems, IEEE/CAA J. Autom. Sin. 9 (2021) 75–89.